\documentclass[11pt]{article}
\usepackage{euscript}
\usepackage{amssymb}
\usepackage{amsfonts}
\usepackage{amsbsy}
\usepackage{epsfig}
\usepackage{amsthm}
\usepackage{amscd}
\usepackage{amstext}
\usepackage{verbatim}
\usepackage{amsmath}
\usepackage{hyperref}
\usepackage{cite}

\usepackage{makecell}
\usepackage{bm}

\usepackage{bbm}
\usepackage{xcolor}
\definecolor{myred}{RGB}{178, 34, 34}
\definecolor{mygreen}{RGB}{34, 139, 34}
\definecolor{myblue}{RGB}{65, 105, 225}
\hypersetup{colorlinks=true,citecolor=blue,urlcolor=black}

\usepackage{amsmath,graphicx,color}
\usepackage{tabularx}
\usepackage{booktabs}
\usepackage{multirow}
\usepackage{graphicx}
\usepackage{subcaption}

\newcolumntype{C}[1]{>{\centering\arraybackslash}m{#1}}

\begin{document}

\begin{center}
{\Large \textbf{Discover physical concepts and equations\\ with machine learning }}

\vspace{1cm}

Bao-Bing Li$^{1}$, Yi Gu$^{2,*}$, and Shao-Feng Wu$^{1,3}$

\vspace{1cm}

$^{1}${\small \textit{Department of Physics, Shanghai University, 200444 Shanghai, China }}\\[0pt]

$^{2}${\small \textit{Department of Physics, Ludwig-Maximilians-Universität, 80333 Munich, Germany}}\\[0pt]

$^{3}${\small \textit{Center for Gravitation and Cosmology, Yangzhou
University, 225009 Yangzhou, China }}

\vspace{0.5cm}

{\small \textit{$^{*}$E-mail:Gu.Yi@campus.lmu.de}}
\end{center}

\vspace{1cm}

\begin{abstract}
Machine learning can uncover physical concepts or physical equations when prior knowledge from the other is available. However, these two aspects are often intertwined  and cannot be discovered independently. We extend \textit{SciNet}, which is a neural network architecture that simulates the human physical reasoning process for physics discovery, by proposing a model that combines Variational Autoencoders (VAE) with Neural Ordinary Differential Equations (Neural ODEs). This allows us to simultaneously discover physical concepts and governing equations from simulated experimental data across various physical systems. We apply the model to several examples inspired by the history of physics, including Copernicus' heliocentrism, Newton's law of gravity, Schrödinger's wave mechanics, and Pauli's spin-magnetic formulation. The results demonstrate that the correct physical theories can emerge in the neural network.
\end{abstract}
\pagebreak

\section{Introduction}

Einstein famously argued that ``It can scarcely be denied that the supreme goal of all theory is to make the irreducible basic elements as simple and as few as possible without having to surrender the adequate representation of a single datum of experience'' \cite{einstein1934method}. The basic elements of physical theories are physical concepts and equations.  Finding simple and elegant physical concepts and equations that can fully describe experimental phenomena has always been the goal of theoretical physicists.

Physics typically advances by building upon existing theories, incorporating new experimental phenomena and mathematical tools to expand them. However, when confronting revolutionary problems, inherited theories may not naturally describe new phenomena. In such cases, breakthrough discoveries are needed to revise the old theories, and this can be a long process. For example, it took approximately 14 centuries from Ptolemy's geocentric system to Copernicus' heliocentric theory, and more than 25 years from Planck's quantum hypothesis to Schrödinger's wave mechanics. Moreover, human research into new phenomena relies on the current level of experimental capability and physical understanding. Subtle physical mechanisms, such as spin, may already be hidden within the data, but they are not easily discovered \cite{maillet2007heisenberg}.

In recent years, the emerging field of AI for Science has developed rapidly. It is being increasingly expected that AI can accelerate scientific discovery and help researchers gain insights unattainable through traditional scientific methods \cite{wang2023scientific}. As a research assistant tool, AI has made significant contributions in various areas of physics, such as particle physics and cosmology, quantum many-body physics, quantum computing, and chemical and material physics \cite{carleo2019machine}. Beyond these successes in specific areas of physics, recent works \cite{wu2019toward,cornelio2023combining,krenn2022scientific,cory2023ai} have demonstrated that AI is capable of conducting autonomous research in general physics. Important physical concepts \cite{iten2020discovering,wang2019emergent,seif2021machine}, physical properties such as symmetries \cite{desai2022symmetry,otto2023unified}, and physical laws such as Newton's laws \cite{lemos2023rediscovering}, Kepler's laws \cite{cory2023ai}, and conservation laws \cite{liu2021machine,alet2021noether,liu2022machine} can all be discovered by AI from data.

Physics, or science, is the attempt at the posterior reconstruction of existence by the process of conceptualization \cite{einstein1940science}. \textit{SciNet}, proposed by Iten et al.~\cite{iten2020discovering}, is an important neural network model that autonomously discovers physical concepts from data without requiring prior knowledge. However, in physics, many concepts and equations are intertwined and hard to discover independently.  Without physical concepts, it is evidently impossible to determine the physical equations. Conversely, without physical equations, we cannot describe the relationships between physical concepts and time, space and themselves, thereby preventing these concepts from being clearly defined. For example, although de Broglie first proposed that matter should possess wave-like properties, the concept of the wave function (later interpreted by Born as a probability amplitude) was introduced by Schrödinger alongside his formulation of the Schrödinger equation \cite{schrodinger1926undulatory}. Current machine learning models mainly focus on discovering either physical concepts or physical equations and often require prior knowledge about the other. In this paper, we will retain \textit{SciNet}'s questioning mechanism but replace its assumption of uniformly evolving latent representations with Neural ODEs \cite{chen2018neural}, an infinite-depth neural network for modeling continuous dynamics. We do not rely on any prior knowledge specific to particular examples, instead making the natural assumption that the evolution of physical concepts is governed by differential equations.  This enhancement renders the framework more general, enabling the simultaneous discovery of physical concepts and governing equations.  Although previous works \cite{choi2022learning,lai2022neural,sholokhov2023physics} have applied similar models in some specific tasks, they mainly focused on the predictive power of neural networks, whereas our goal is to extract meaningful physical information from the latent representations. We are particularly concerned with whether important concepts and equations in the history of theoretical physics can be discovered. In view of these, we apply our model to four examples in physics: Copernicus' heliocentrism, Newton's law of gravity, Schrödinger's wave mechanics, and Pauli's spin-magnetic formulation. Despite the redundancy in the latent space, we find that AI can discover correct physical theories consistent with those in textbooks.

The first three examples refer to previous works \cite{iten2020discovering,daniels2015automated,wang2019emergent}. Compared to their models, which focus on identifying either concepts or equations, or are applied to specific systems, our model can simultaneously discover both across different systems. In the case of Copernicus' heliocentrism, \textit{SciNet} utilized an implicit questioning mechanism (e.g.,~``Where is the particle at time \(t\)?'') by providing sequential observed data \cite{iten2020discovering}. We use the same mechanism in the first example. However, in the following three examples, an explicit questioning mechanism regarding control variables such as potential fields is used (e.g.,~“Where is the particle at time \(t\) under a potential field \(V\)?”). This allows the model to handle systems with additional control variables, thereby extending its applicability. 

The rest of the paper will be arranged as follows. Section \ref{app:method} introduces our neural network architecture that mimics human physical modeling. In 
Section \ref{app:result}, we describe the four physical examples and exhibit the training results. Conclusion and discussion are presented in Section \ref{app:conclusion and discussion}. Additionally, there are five appendices. Appendix \ref{app:Appendix A} provides the details of the neural network architecture and the training process. Appendix \ref{app:Appendix B} provides a brief introduction to VAE and explains how the Kullback–Leibler (KL) divergence promotes disentangled latent representations. Appendix \ref{app:Appendix C} generates random potential functions. In Appendix \ref{app:Appendix D}, we introduce the Pauli equation for silver atoms in a uniform and weak magnetic field. In Appendix \ref{app:Appendix E}, we attempt to uncover independent concepts by assuming that the true equations are expressible as second-order equations.

\section{Method}\label{app:method}

The development of physical theories begins with the observation of natural phenomena. Physicists analyze the observed data to propose physical concepts such as particles, waves, and fields, and formulate equations to explain the observed phenomena. Throughout this process, physicists refine theoretical models by posing questions, such as, ``What would happen if we altered a specific condition?'' and seeking answers. When experimental results support the theory, it is reinforced; when they contradict the predictions, the theory may need to be revised. In the following, we will develop a general machine learning-based method for physics research. We will model our approach on the way human physicists construct physical theories. Our specific goal is to enable AI to automatically discover physical concepts and their corresponding differential equations.

\textbf{Discovering physical concepts.}—Suppose there is a connection between the direct observed data and the underlying physical concepts. We use a VAE \cite{kingma2013auto} to learn the mapping and the inverse mapping between them\footnote{Strictly speaking, we use a variant of the standard VAE, which will be specified later in the text.}. A standard VAE consists of a probabilistic encoder and a probabilistic decoder. The probabilistic encoder maps the input data space to a latent space, producing latent representations. The probabilistic decoder then uses these latent representations to reconstruct the input data or generate new data. In physics, physical concepts serve as a simple and essential representation of the physical observed data. When we input observed data into the VAE, we expect the latent space to store information corresponding to the underlying physical concepts. To encourage independence among latent variables, we adopt the \(\beta\)-VAE loss function \cite{higgins2017beta}, which is an extension of the standard VAE loss function \cite{kingma2013auto}. This loss function comprises the reconstruction error, the KL divergence, and a hyperparameter \(\beta\) that modulates the strength of the KL divergence. Previous studies \cite{iten2020discovering,higgins2017beta,frohnert2024explainable,fernandez2024learning} have demonstrated that introducing the KL divergence as a regularization term and varying its strength encourages the latent space to learn minimal, sufficient, and uncorrelated representations. Based on this insight, we incorporate the KL divergence into our architecture and tune the hyperparameter \(\beta\) to optimize the latent space representations, bringing it closer to the requirements of physics.

\textbf{Discovering physical equations.}—The physical equations that govern the evolution of physical concepts are another essential element of theory. Several machine learning techniques have been developed to identify the underlying governing equations from sequential data, such as Recurrent Neural Networks (RNNs) \cite{cho2014learning}, SINDy \cite{brunton2016discovering}, and Koopman theory \cite{brunton2021modern}. In this work, we utilize Neural ODEs \cite{chen2018neural} to discover physical equations. Neural ODEs are a class of machine learning models that integrate traditional neural networks with ODEs in a natural way. Unlike discrete-time models such as RNNs, Neural ODEs can model continuous-time dynamics with high memory efficiency. Neural ODEs do not require predefined candidate functions and can directly learn complex nonlinear dynamics from data, whereas Koopman theory may be limited in its ability to linearize nonlinear systems, and SINDy may face challenges when the nonlinear complexity of the system exceeds the expressive capacity of the candidate functions. In view of these and other advantages, Neural ODEs have found many applications in different fields of physics, see e.g. \cite{hashimoto2021neural,rrapaj2021inference,chen2022forecasting,chen2022learning,zhao2023optical,li2023metalearning,liu2024application,gu2025neural}.

Consider that the governing equations of dynamical systems can be represented by an ODE, in the general form:
\begin{equation}
\frac{dh(t)}{dt} = f(t, h(t);\zeta) ,
\end{equation}
where \( t \) is time, \( h(t) \) are state variables, and \( f \) can be referred to the governing functions of Neural ODEs, typically represented by a neural network with the parameter $\zeta$.

Given an initial value for the state variable \( h(t_0) \) and a neural network \(f\), Neural ODEs use a numerical ODE solver to generate the state variable at any moment through forward propagation: 
\begin{equation}
{h}(t_1) = h(t_0) + \int_{t_0}^{t_1} {f}(t, {h}(t);\zeta) \, dt.
\end{equation}

The backpropagation of Neural ODEs is implemented using the adjoint sensitivity method \cite{chen2018neural}. Suppose there is a loss function dependent on the output of the ODE solver:
\begin{equation}
L(h(t_1)) = L(h(t_0) + \int_{t_0}^{t_1} {f}(t, h(t); \zeta) \, dt).
\end{equation}
The adjoint state is defined as:
\begin{equation}
    a(t) = \frac{\partial L}{\partial h(t)}.
\end{equation}
It has been proven that the adjoint dynamics are governed by another ODE \cite{chen2018neural}
\begin{equation}
    \frac{da(t)}{dt} = -a(t) \cdot \frac{\partial {f}}{\partial h}.
\end{equation}
The parameter gradient can be calculated using the integration
\begin{equation}
    \frac{\partial L}{\partial \zeta} = \int_{t_1}^{t_0} a(t) \cdot \frac{\partial {f}}{\partial \zeta} dt.
\end{equation}

\begin{figure}[!b]
\centering
\captionsetup{font=footnotesize, labelfont=bf}
\includegraphics[width=15cm]{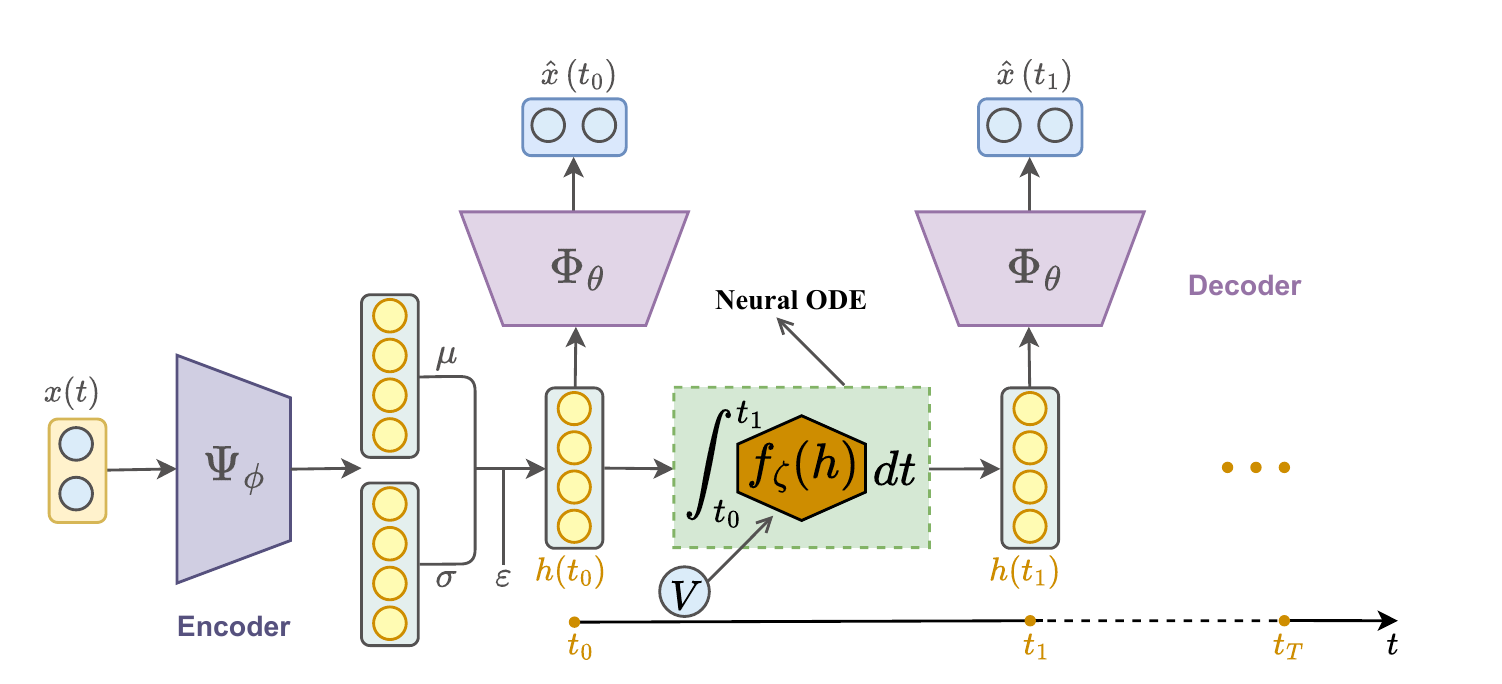}
\caption{ Neural network architecture. Our model consists of two components: the VAE and the Neural ODE. The observed data \( x(t) \) are processed by the encoder \( \Psi_\phi \), which outputs the distribution parameters \( \mu \) and \( \sigma \). Then, an initial latent state \( h(t_0) \) is sampled using the reparameterization trick as \( h(t_0) = \mu + \sigma\cdot \varepsilon \), where $\varepsilon$ is an auxiliary parameter. The latent state \( h(t_0) \) is used as the initial condition of the Neural ODE. The Neural ODE is characterized by the governing function \(f\), which is a neural network with parameter \(\zeta\) that depends on the state \(h(t)\) and an external control variable \(V\). Using a numerical ODE solver, the Neural ODE outputs a series of latent states at different moments \( h(t_i) \). Using these latent states the decoder \( \Phi_\theta \) reconstructs the corresponding observed data \( \hat{x}(t_i) \). }
\label{fig:Neural Network Architecture}
\end{figure}
\textbf{Neural network architecture.}—The overall neural network architecture is illustrated in Fig.~\ref{fig:Neural Network Architecture}. We begin by collecting the time-sequential observed data in the observable space, denoted as 
\begin{equation}
{x}(t) = \left\{ x^{(k)}_{j}(t_i) \;\middle|\; i = 0, \dots, I{-}1,\; j = 1, \dots, J,\; k = 1, \dots, K \right\},
\end{equation}
where \( x^{(k)}_{j}(t_i) \) denotes the \( j \)-th observed variable at the \( i \)-th moment of the \( k \)-th data sequence. According to different tasks and model’s performance, the complete time-sequential  data or only the data at the first moment (uniformly denoted as \(x(t)\) for simplicity) are input into the probabilistic encoder \(\Psi_\phi\). The encoder \(\Psi_\phi\) outputs a probability distribution in the latent space, represented by the mean \(\mu\) and standard deviation \(\sigma\):
\begin{equation}
({\mu}, {\sigma}) = \Psi_\phi({x}(t)).
\end{equation}

Next, we sample from the distribution to obtain the latent state at the first moment:
\begin{equation}
{h}(t_0) \sim \mathcal{N}({\mu}, {\sigma}).
\end{equation}
Here, \({h}\) is a vector whose dimension is the same as that of the latent space. In a standard VAE, the reparameterization trick is commonly employed, where the latent state is generated by \( h = \mu + \sigma\cdot \varepsilon \) and the auxiliary parameter \(\varepsilon\) is typically drawn from  a standard normal distribution. This is equivalent to injecting noise into the latent space. However, our experiments indicate that large noise introduced by \(\varepsilon\), when coupled with a dynamical system, leads to unstable training and increased reconstruction error. To address this, we set \(\varepsilon = 0\), thereby making the latent-space sampling process deterministic. Although this choice departs from the conventional VAE framework, we retain the standard deviation \(\sigma\) to calculate the KL divergence, resulting in a VAE variant that alleviates the adverse effects of high noise on dynamics inference while preserving the partial benefit of modeling uncertainty. For simplicity, we still refer to this variant as VAE throughout the paper.

We use Neural ODEs to define the dynamics of the latent state through a differential equation to be learned:
\begin{equation}
\frac{d{h}(t)}{dt} = {f}(t, {h}(t), V; \zeta).
\end{equation}
Here, the neural network \(f\) takes time \(t\), the state variable \({h}(t)\), and an optional control variable \(V\).

We treat \({h}(t_0)\) as the initial condition for Neural ODEs and employ a numerical solver to compute the latent states at the subsequent moments. The decoder \(\Phi_\theta\) then takes the latent states \({h}(t)\) and reconstructs the corresponding observed data:
\begin{equation}
\hat{x}(t)= \left\{ \hat{x}^{(k)}_{j}(t_i) \;\middle|\; i = 0, \dots, I{-}1,\; j = 1, \dots, J,\; k = 1, \dots, K \right\}.
\end{equation}

The encoder \(\Psi_\phi\), decoder \(\Phi_\theta\), and the governing functions \(f\) in our model are all fully connected networks consisting of multiple hidden layers and nonlinear activation functions. Additional details can be found in Appendix \ref{app:Appendix A}.

\textbf{Loss function.}—We quantify the discrepancy between the physical theories discovered by the machine and the actual physical theories by calculating the error between the reconstructed data \( {\hat{x}}(t)\) and the true observed data \( {x}(t) \). This process is analogous to how physicists refine theories based on comparative experimental results. Using the gradient descent algorithm, the neural network model iteratively updates the parameters representing the learned theory, minimizing the reconstruction error to ultimately identify a physical theory that adequately describes the observed phenomena. We use the squared \(L2\) norm to measure reconstruction error, incorporating the KL divergence to encourage independence in the latent space. The resulting loss function is given by\footnote{In the second example (see Section \ref{app:example2}), we also include the Mean Relative Error between \({x}(t) \) and \({\hat{x}}(t)\) as a regularization term. This step improves the accuracy of the learned concepts and equations in that specific example, but does not yield a noticeable improvement in the other cases.}:
\begin{equation}
L({x}; \phi, \zeta, \theta) 
= \frac{1}{I J}\sum_{j=1}^{J}\sum_{i=0}^{I-1}\left\|{x}_{j}(t_i)-\hat{x}_{j}(t_i)\right\|^{2}_2
 + \beta \cdot D_{\mathrm{KL}}\bigl[p_{{\phi}}({h}|{x}) \,\|\, p({h})\bigr].
\end{equation}
The first term represents the reconstruction error, where \({x}_{j}(t_i) = \{x^{(k)}_{j}(t_i) \mid k = 1, \dots, K\}\) is understood as a vector. The second term is the KL divergence, and its detailed expression is provided in Appendix \ref{app:Appendix B}. There, we also explain how the KL divergence promotes independence among the latent variables.

\textbf{Number of physical concepts.}—Another key hyperparameter in our model is the dimension of the latent space, which corresponds to the number of physical concepts required to describe the phenomena. To ensure objectivity, we do not assume prior knowledge of this parameter but treat it as a hyperparameter to be determined by the ablation experiments. We train models with different latent space dimensions under the same conditions and plot the corresponding loss function curves. According to Occam's Razor, the optimal number of physical concepts should fully describe the observed data using the smallest possible number. Therefore, we select the smallest latent space dimension that does not result in a significant increase in the loss function as the optimal number of physical concepts.

\section{Results}\label{app:result}

\subsection{Copernicus' heliocentrism}
\begin{figure}[!t]
    \centering
    \captionsetup{font=footnotesize, labelfont=bf}
    \setlength{\abovecaptionskip}{0pt}
    \setlength{\belowcaptionskip}{0pt}

    \begin{minipage}[b]{0.40\textwidth}
        \centering
        \includegraphics[width=\textwidth]{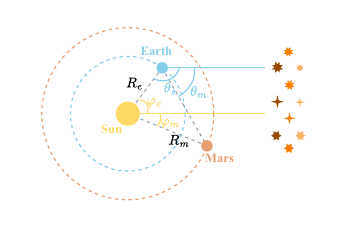}
        \subcaption{}
    \end{minipage}
    \hspace{3em}
    \begin{minipage}[b]{0.40\textwidth}
        \centering
        \includegraphics[width=\textwidth]{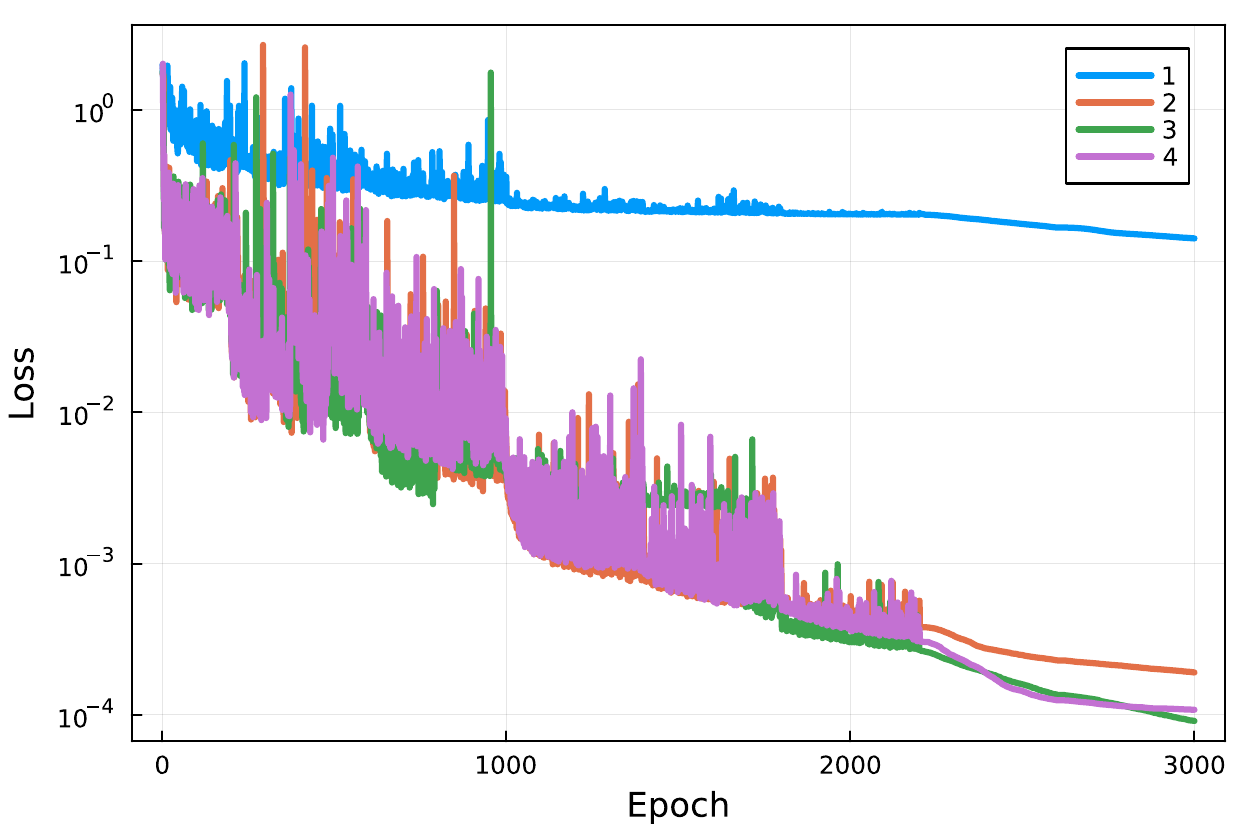}
        \subcaption{}
    \end{minipage}

    \vspace{5pt}

    \begin{minipage}[b]{0.53\textwidth}
        \centering
        \begin{minipage}[b]{0.31\textwidth}
            \centering
            \includegraphics[width=\textwidth]{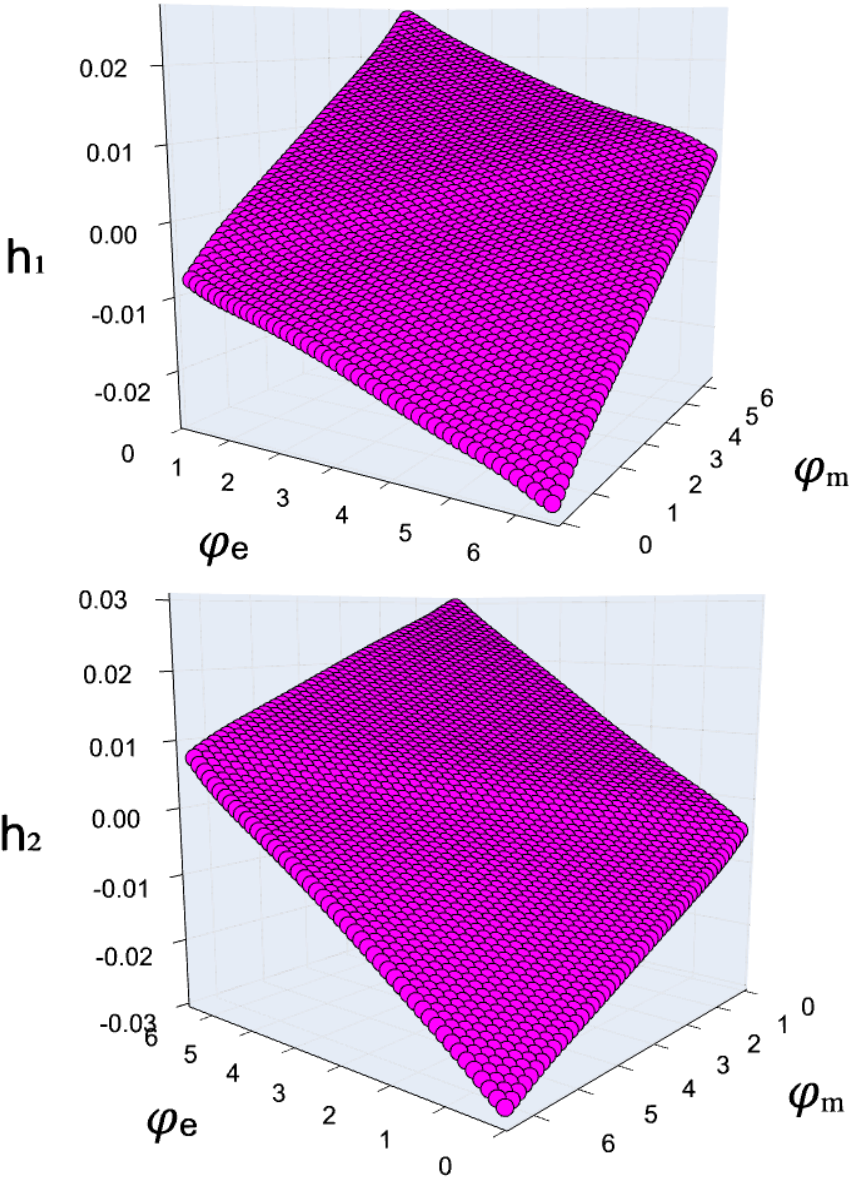}
            \subcaption*{(1)}
        \end{minipage}\hspace{0.05em}
        \begin{minipage}[b]{0.31\textwidth}
            \centering
            \includegraphics[width=\textwidth]{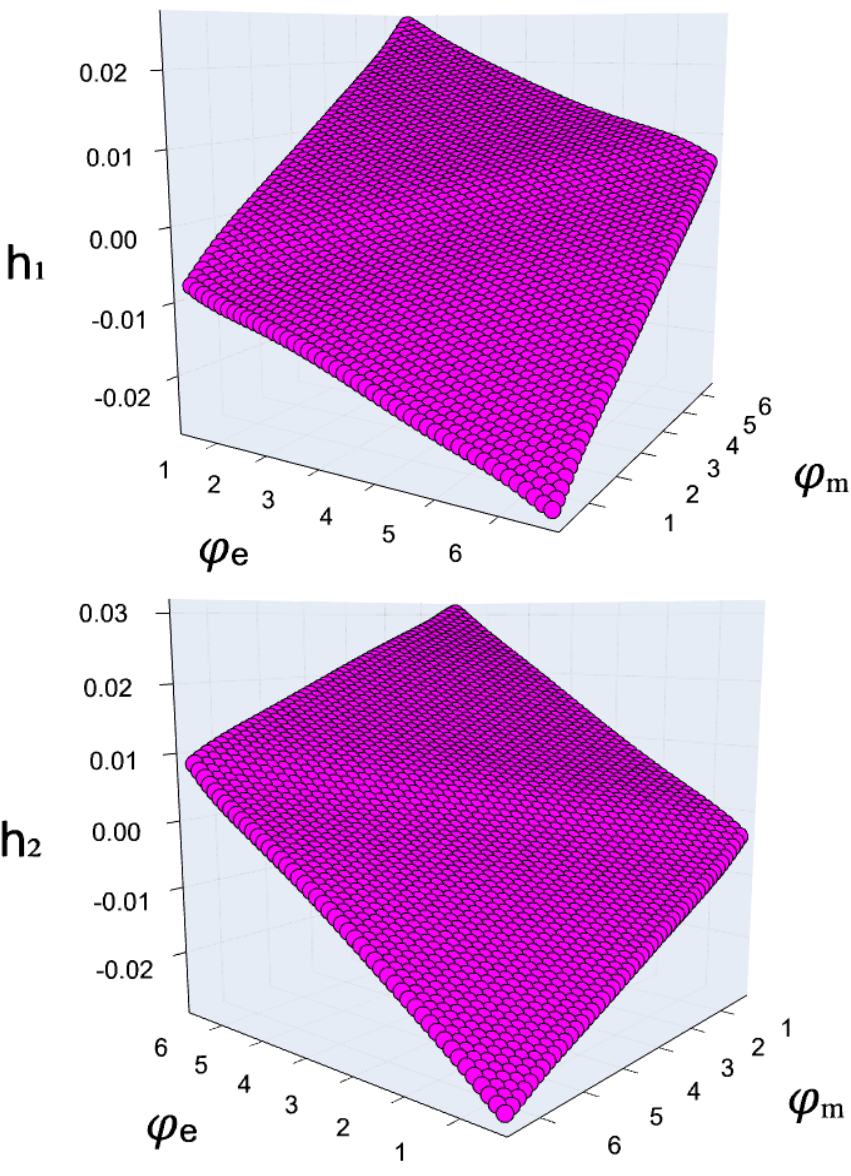}
            \subcaption*{(2)}
        \end{minipage}\hspace{0.05em}
        \begin{minipage}[b]{0.31\textwidth}
            \centering
            \includegraphics[width=\textwidth]{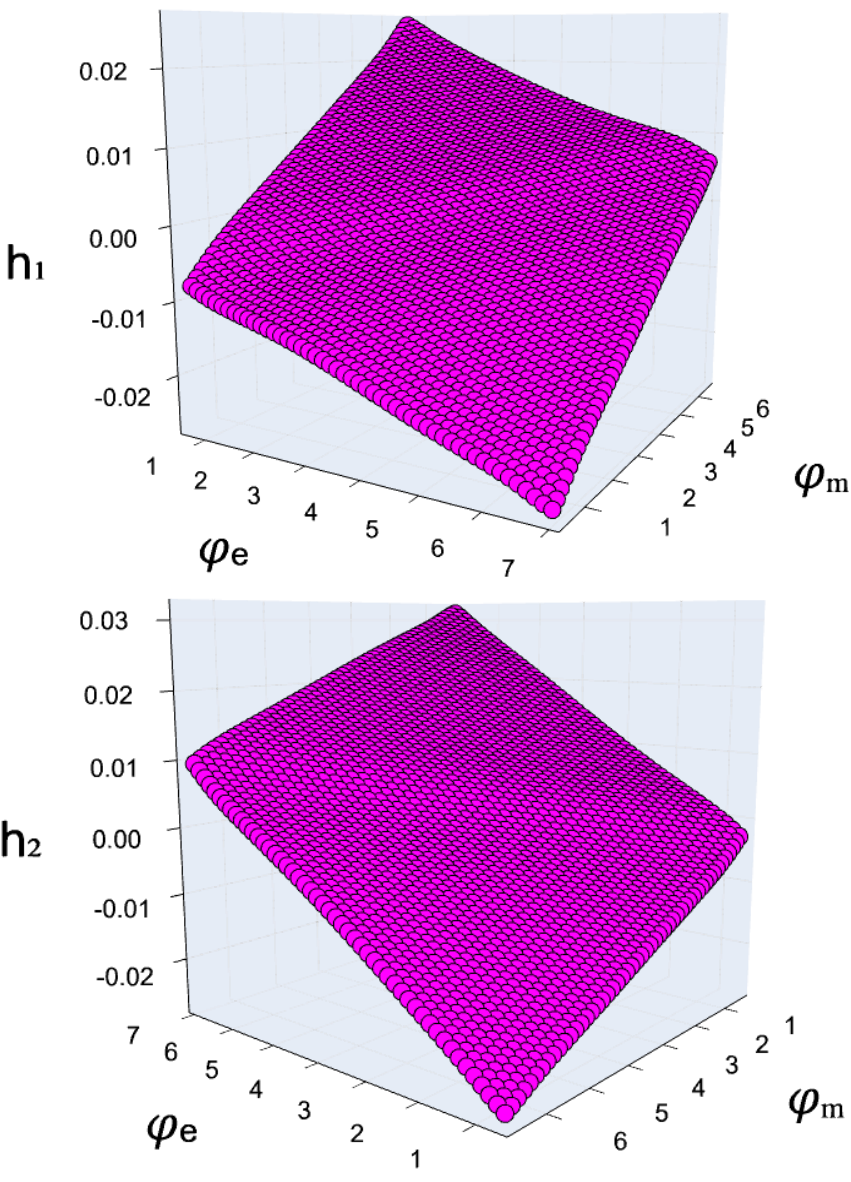}
            \subcaption*{(3)}
        \end{minipage}
        \subcaption{}
    \end{minipage}
     \hspace{1em}
    \begin{minipage}[b]{0.4\textwidth}
        \centering
        \includegraphics[width=\textwidth]{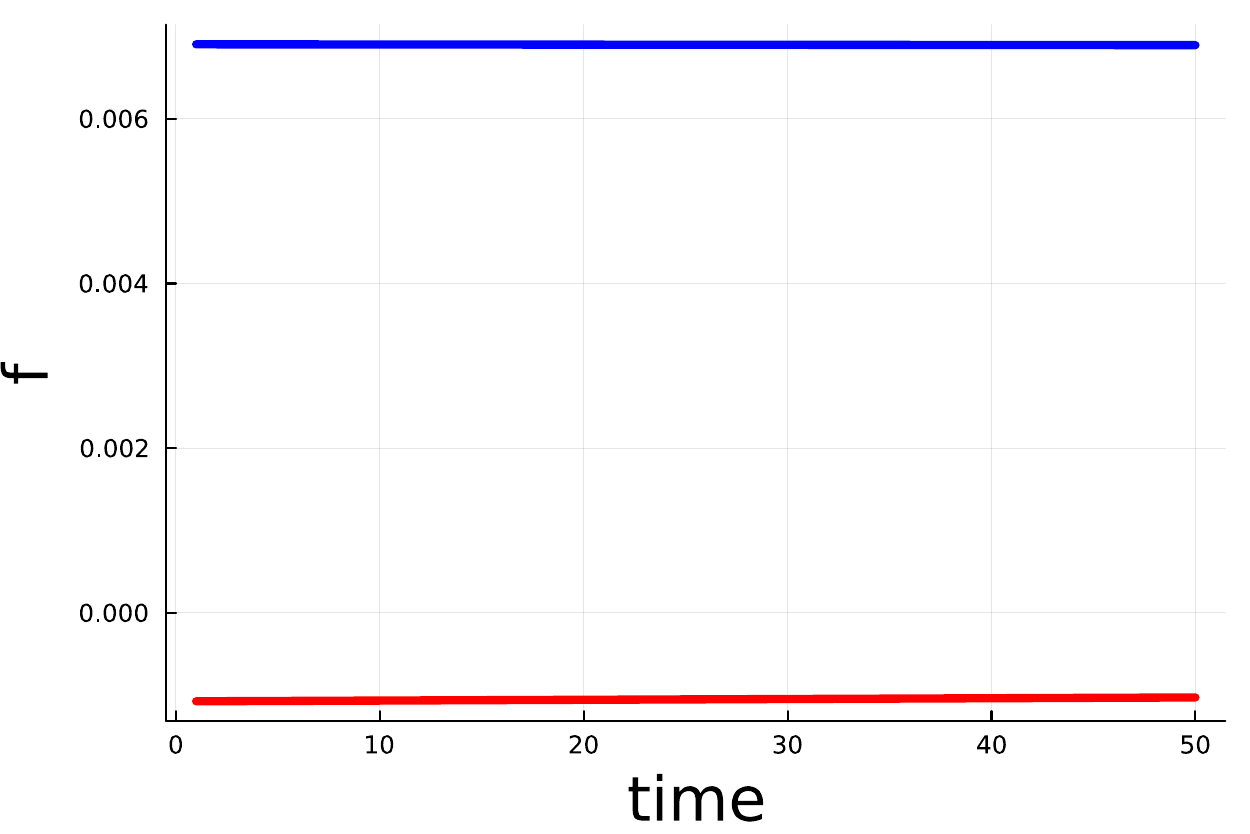}
        \subcaption{}
    \end{minipage}

    \caption{Heliocentric solar system. (a) The \(\varphi_e\) and \(\varphi_m\) denote the heliocentric angles of Earth and Mars, while \(\theta_s\) and \(\theta_m\) represent the geocentric angles of the Sun and Mars. These angles are measured against a background of fixed stars. Both the Earth and Mars move in uniform circular motion around the Sun. (b) Impact of latent dimensions on loss, showing a two-dimensional space as optimal. (c) The latent representations capture linear combinations of heliocentric angles at the 10th, 25th, and 40th moments, corresponding to plots (1), (2), and (3), respectively. (d) The governing functions \(f\) learned indicate constant rates of change for two latent representations.}
    \label{fig:Heliocentric Model of the Solar System}
\end{figure}

Copernicus' heliocentric theory played a revolutionary role in the history of science and profoundly impacted our understanding of the universe. It also signifies a general research paradigm in physics that starts from observed data and seeks to discover physical laws through simplicity and predictability. Here, we consider the heliocentric model where Mars and Earth orbit the Sun in uniform circular motion, as shown in Fig.~\ref{fig:Heliocentric Model of the Solar System}(a). The key question then arises: Can AI, without any prior knowledge, replicate the work of Copernicus?

As shown in Fig.~\ref{fig:Heliocentric Model of the Solar System}(a), the distance between Earth and Mars is given by:
\begin{equation}
    d = \sqrt{R_e^2 + R_m^2 - 2 \cdot R_e \cdot R_m \cdot \cos(\varphi_e + \varphi_m)},
\end{equation}
where \(R_e\) is the distance from the Earth to the Sun, and \(R_m\) is the distance from Mars to the Sun. With distant fixed stars as the reference frame, \(\varphi_e\) and \(\varphi_m\) are the angles of Earth and Mars relative to the Sun, respectively. Then, the angles of the Sun and Mars relative to the Earth, \(\theta_s\) and \(\theta_m\), are given by:
\begin{equation}
   \theta_s = \pi - \varphi_e,
\end{equation}
\begin{equation}
   \theta_m = atan2\left({\sin(\theta_m)},{\cos(\theta_m)}\right),
\end{equation}
where
\begin{equation}
    \sin(\theta_m) = \frac{R_e \sin(\varphi_e) + R_m \sin(\varphi_m)}{d},
\end{equation}
\begin{equation}
   \cos(\theta_m) = \frac{R_m \cos(\varphi_m) - R_e \cos(\varphi_e)}{d}.
\end{equation}

In Ref. \cite{iten2020discovering}, Iten et al.~used \textit{SciNet} to investigate whether the neural network can autonomously discover the concept of heliocentric angles from time-evolving geocentric angle data, provided that the hidden physical concepts evolve uniformly over time. Although uniform motion is a natural and simple assumption, we hope to relax it and only suppose that the latent variables evolve through differential equations.

Our physical setup is the same as in \cite{iten2020discovering}. The training dataset consists of randomly selected subsequences of geocentric angles \(\theta_m(t)\) and \(\theta_s(t)\) recorded weekly throughout Copernicus's entire lifespan, totaling 3665 observations. The neural network's input data are the geocentric angles at the first moment \(\theta_s(t_0)\) and \(\theta_m(t_0)\), and the label data is the sequence of the geocentric angles across 50 moments.

To proceed, we consider that a non-autonomous ODE can always be transformed into an autonomous ODE by increasing the dimension of the system \cite{zhi2022learning}. Putting this fact together with our scheme that the dimension of the latent space will be determined by the ablation experiments, we can set the differential equations to be discovered as the autonomous form without loss of generality:
\begin{equation}
    \frac{d{h}(t)}{dt} = {f}({h}(t);\zeta),
\end{equation}
where the time \(t\) is not directly input into the governing functions \(f\). 

Ablation experiments show, as depicted in Fig.~\ref{fig:Heliocentric Model of the Solar System}(b), that although the loss is slightly lower in three-dimensional and four-dimensional latent spaces compared to the two-dimensional space, the two-dimensional space maintains the minimum representation necessary to describe the physical system with its much lower loss than one-dimensional space. Therefore, we infer the optimal latent dimension for the system to be two. Fig.~\ref{fig:Heliocentric Model of the Solar System}(c) displays the two latent representations which store linear combinations of the true heliocentric angles \(\varphi_e\) and \(\varphi_m\) at the 10th, 25th, and 40th moments. This indicates that the neural network has discovered the linear combinations of two heliocentric angles as the optimal representations to infer the observed geocentric angles. Fig.~\ref{fig:Heliocentric Model of the Solar System}(d) shows that the learned governing functions \({f}({h}(t);\zeta)\) in Neural ODEs are two constants, indicating that the two latent representations are evolving at a constant speed, which is consistent with the actual setting.

\subsection{Newton's law of gravity }\label{app:example2}
\begin{figure}[!b]
    \centering
    \captionsetup{font=footnotesize, labelfont=bf}
    \setlength{\abovecaptionskip}{0pt}
    \setlength{\belowcaptionskip}{0pt}

    \begin{minipage}[b]{0.2\textwidth}
        \centering
        \includegraphics[width=\textwidth]{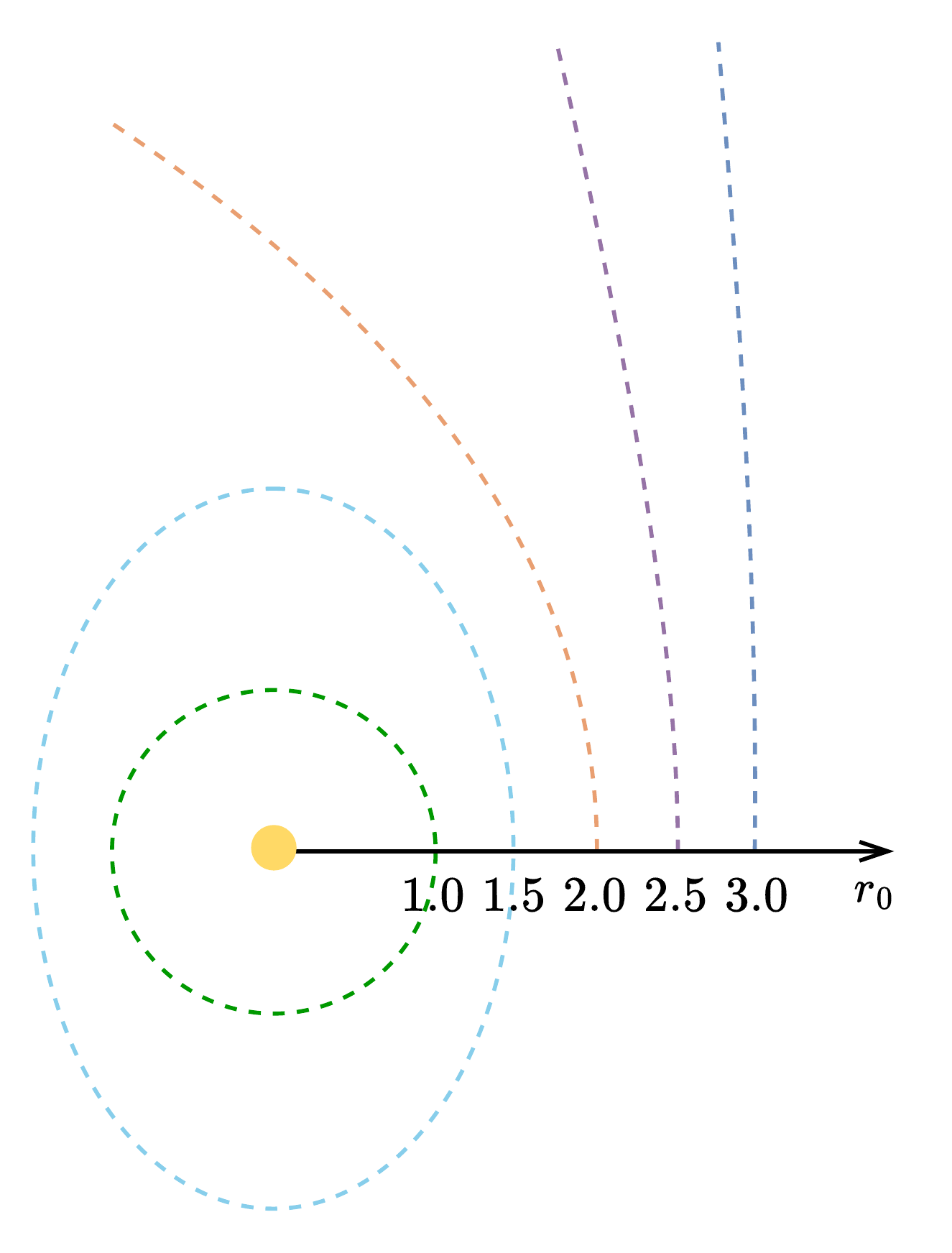}
        \subcaption{}
    \end{minipage}\hfill
    \begin{minipage}[b]{0.39\textwidth}
        \centering
        \includegraphics[width=\textwidth]{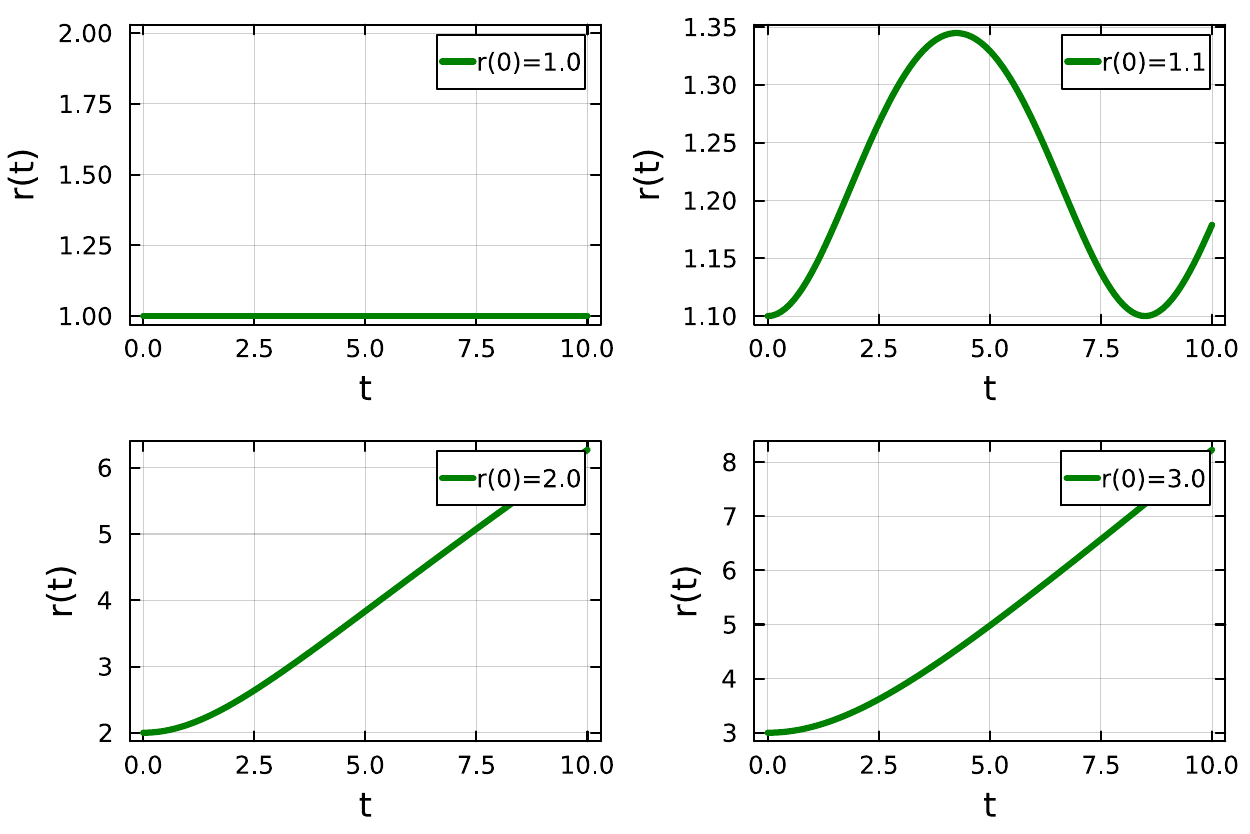}
        \subcaption{}
    \end{minipage}\hfill
    \begin{minipage}[b]{0.39\textwidth}
        \centering
        \includegraphics[width=\textwidth]{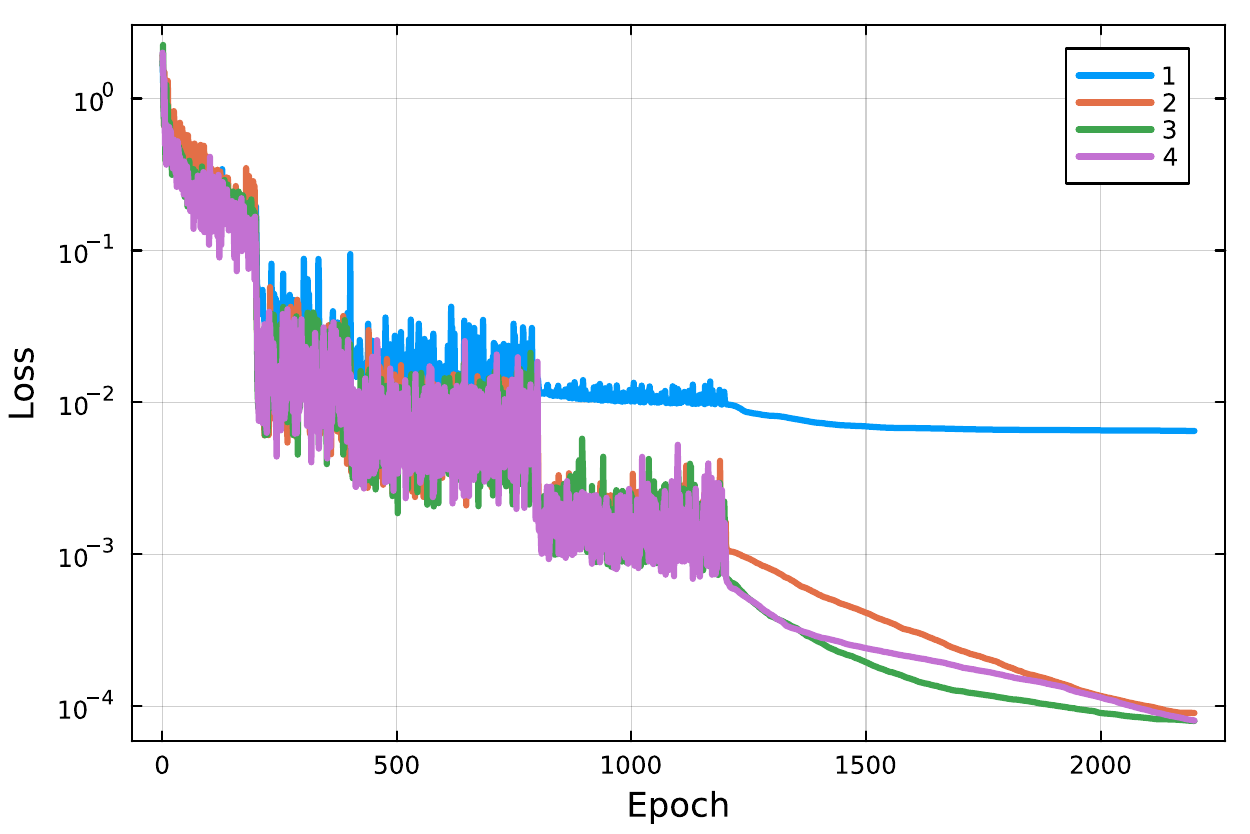}
        \subcaption{}
    \end{minipage}

    \vspace{10pt}

    \begin{minipage}[b]{0.48\textwidth}
        \centering
        \includegraphics[width=\textwidth]{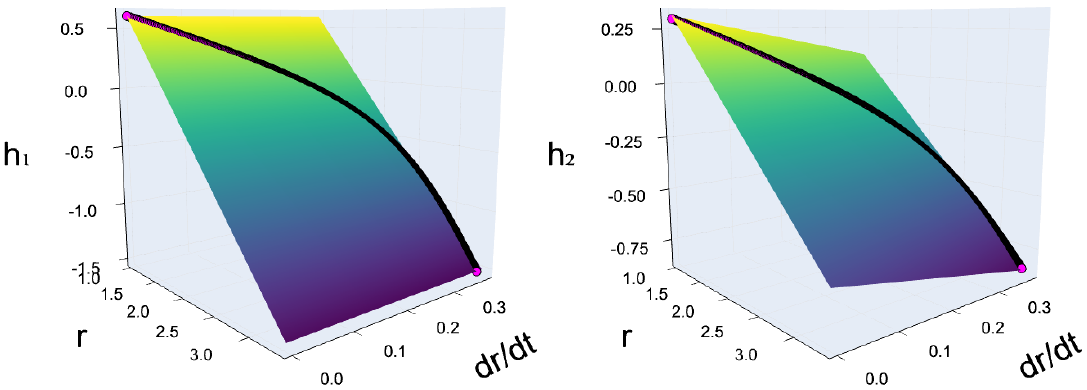}
        \subcaption*{(1)}
        \vspace{5pt}

        \includegraphics[width=\textwidth]{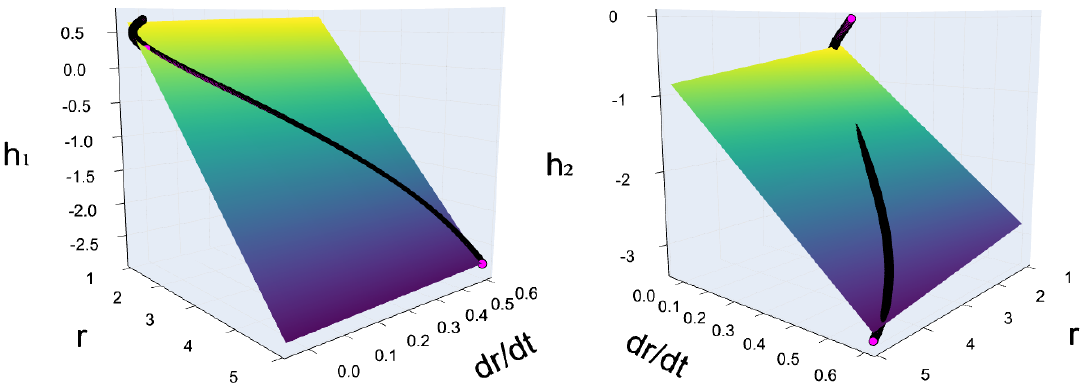}
        \subcaption*{(2)}
        \vspace{5pt}

        \includegraphics[width=\textwidth]{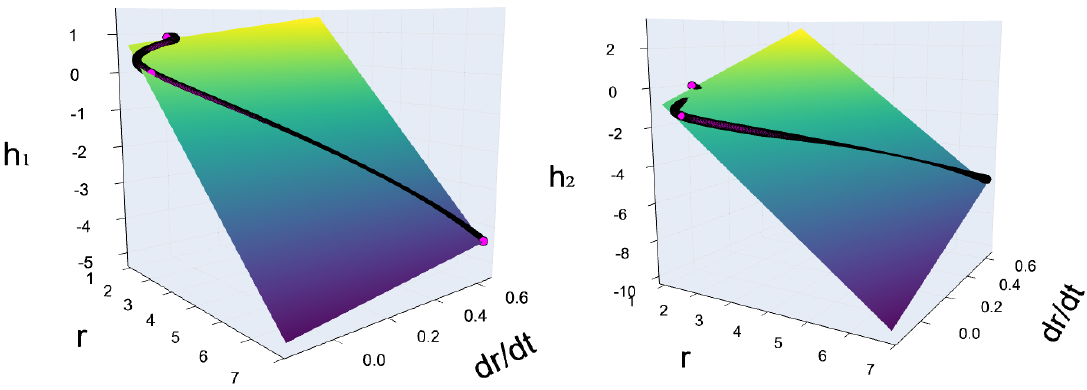}
        \subcaption*{(3)}
        \subcaption{}
    \end{minipage}\hfill
    \begin{minipage}[b]{0.45\textwidth}
        \centering
        \includegraphics[width=\textwidth]{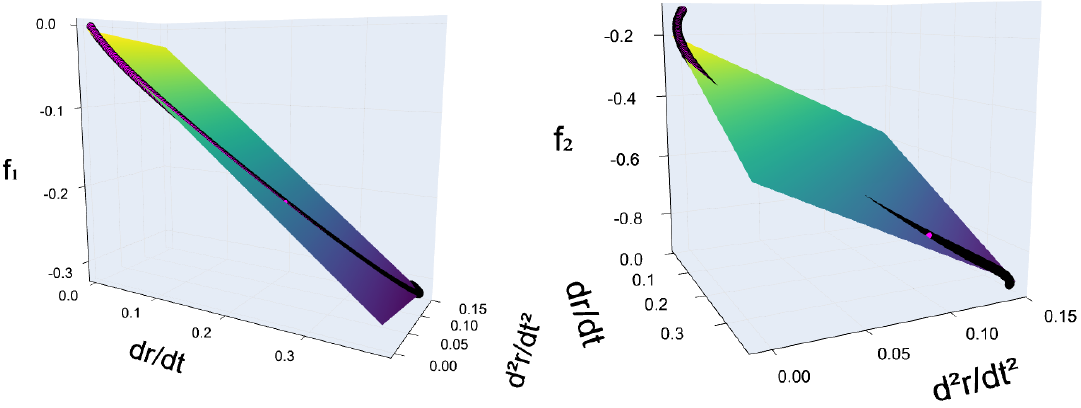}
        \subcaption*{(1)}
        \vspace{5pt}

        \includegraphics[width=\textwidth]{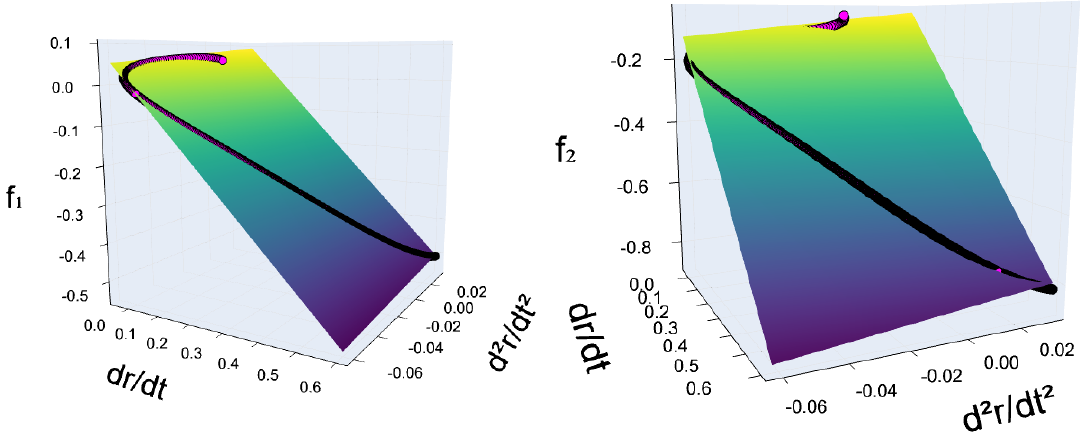}
        \subcaption*{(2)}
        \vspace{5pt}

        \includegraphics[width=\textwidth]{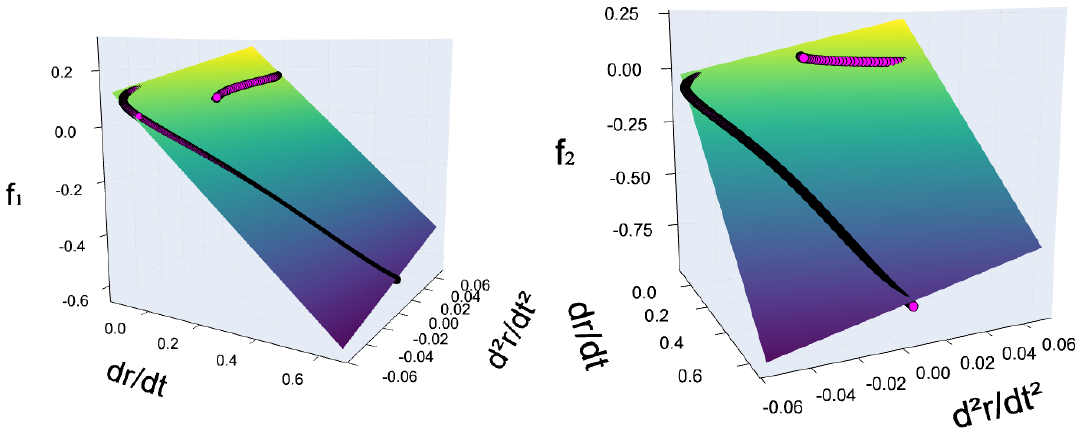}
        \subcaption*{(3)}
        \subcaption{}
    \end{minipage}

    \caption{The law of gravity.
    (a) Different types of motion trajectories determined by \( r_0 \).  
    (b) Distances \( r(t) \) from the central object with different \( r_0 \).  
    (c) Impact of latent dimensions on loss, showing that a two-dimensional space is optimal.  
    (d) The latent representations store linear combinations of \( r \) and \( dr/dt \) at the 20th (1), 50th (2), and 80th (3) moments.   
    (e) The governing functions learned by Neural ODEs are linear combinations of \( {dr}/{dt} \) and \( {d^2 r}/{dt^2} \) at the 20th (1), 50th (2), and 80th (3) moments. The \( {d^2 r}/{dt^2} \) should be understood as the right hand of Eq.~(\ref{gravitation equation}). In (d) and (e), the learned latent representations and governing functions appear as winding curves, which lie almost entirely on the reference plane obtained through linear fitting.}
    \label{fig:Model of Universal Gravitation}
\end{figure}

The law of universal gravitation explicitly defines a fundamental characteristic of our universe: gravity. Newton derived this law based on empirical observations of trajectories of the planets, the Moon, and the apocryphal apple. In \cite{daniels2015automated}, an adaptive method called \textit{Sir Isaac }was developed, which correctly infers the phase space structure for planetary motion. While this method has shown significant advantages in dealing with two classes of nonlinear systems, it is worth noting that the hidden variable learned to model the law of gravity is defined up to a power \cite{daniels2015automated}. Here we do not impose any restrictions on the equations and hope that the hidden variables can be identified more exactly. We employ the same physical setup as in \cite{daniels2015automated}, simulating an object with mass  \(m\) moving under a gravitational field by another central object with mass \(M \) where \(M \gg m\). The evolution of the distance \( r(t) \) between them follows the dynamic equation:
\begin{equation}
     \frac{d^2 r}{dt^2} = \frac{h^2}{r^3} - \frac{GM}{r^2},
\end{equation}
where \( h = ({v}_0 \cdot \hat{\theta}) r_0 \) is the specific angular momentum, \( {v}_0 \) is the initial velocity, \( r_0 \) is the initial distance, \( \hat{\theta} \) is a unit vector perpendicular to the line connecting the two objects, and \( G \) is the gravitational constant. Setting the initial velocity parallel to \( \hat{\theta} \), \( {GM}/{v_0^2} \) as the unit of distance and \({GM}/{v_0^3} \) as the unit of time, the dynamic equation simplifies to:
\begin{equation}
    \frac{d^2 r}{dt^2} = \frac{1}{r^2} \left(\frac{r_0^2}{r} - 1\right).
    \label{gravitation equation}
\end{equation}
The motion trajectories of the object with mass $m$ are shown in Fig.~\ref{fig:Model of Universal Gravitation}(a). During the generation of training data, \( r_0 \) is uniformly sampled 1000 times between 1 and 3, covering all possible trajectory types: circular when \( r_0 = 1 \); elliptical when \( 1 < r_0 < 2 \); parabolic when \( r_0 = 2 \); and hyperbolic when \( r_0 > 2 \).

Initial distances \( r_0 \)  serve as input data and control variables in the model to explore how the trajectories of objects change with different initial distances \( r_0 \). The label data are distances \( r(t) \)  observed over the sequential 100 moments, as shown in Fig.~\ref{fig:Model of Universal Gravitation}(b). The differential equations to be discovered are set as follows:
\begin{equation}
    \frac{d{h}(t)}{dt} = {f}({h}(t), r_0;\zeta)\label{22},
\end{equation}
where \(r_0\) is a control variable.

Ablation experiments show, as displayed in Fig.~\ref{fig:Model of Universal Gravitation}(c), that the optimal number of physical concepts required to describe this physical system is two. In Fig.~\ref{fig:Model of Universal Gravitation}(d), we employ a planar ansatz to fit the latent representations learned by the neural network at the 20th, 50th, and 80th moments. The ansatz is given by 
\begin{equation}
    h_j(t) = a_j \cdot r(t) + b_j\cdot \frac{dr(t)}{dt}+ c_j.
\end{equation}
Here, \( j\) takes the values 1 and 2, and \(a_j\), \(b_j\), and \(c_j\) are fitting constants. The fitting result suggests that the neural network has identified the linear combinations of the distance \( r \) and the velocity \( dr/dt \) as the optimal concepts for describing this gravitational system. Next, we want to see what Neural ODEs have learned. We calculate the derivatives of two latent representations with respect to time \( t \):
\begin{equation}
    f_j(t)=\frac{dh_j(t)}{dt} = a_j\cdot \frac{dr(t)}{dt}+ b_j\cdot\frac{d^2 r(t)}{dt^2}= a_j\cdot \frac{dr(t)}{dt}+ b_j\cdot\left[\frac{1}{r^2} \left(\frac{r_0^2}{r} - 1\right)\right] \label{dh1},
\end{equation}
where we have introduced the concepts of the learned linear combinations form into the differential equations. One can see that the governing functions \(f\) in Neural ODEs are expected to output linear combinations of \( {dr}/{dt} \) and \( {d^2 r}/{dt^2} \), where \( {d^2 r}/{dt^2} \) should be understood as the right hand of Eq.~(\ref{gravitation equation}). Fig.~\ref{fig:Model of Universal Gravitation}(e) shows the governing functions learned by the Neural ODEs at the 20th, 50th, and 80th moments, indicating that our model has actually captured the correct dynamics.

\subsection{Schrödinger's wave mechanics}
Schrödinger's wave mechanics described particles as quantum waves and enabled precise predictions of atomic behavior. Its unification with Heisenberg’s matrix mechanics and the introduction of Born's probabilistic explanation reshaped modern physics and philosophical views on reality. In~\cite{wang2019emergent}, Wang et al. proposed an intriguing introspective learning architecture, called ``Schrödinger machine", which includes a translator and a knowledge distiller. By using a two-stage method that involves Taylor expansion and linear projection to update its hidden states and generate outputs, it can extract the concept of quantum wave functions and the Schrödinger equation. Here, we aim to use our general model to achieve the same task.

In this regard, we adopt the same physical setting as in \cite{wang2019emergent}, considering a single quantum particle moving in one-dimensional space under random potential functions\footnote{Note that physicists can collect density distributions of Bose-Einstein condensates in potential traps of various shapes through cold atom experiments\cite{lye2005bose}.}. Accordingly, we write down the Schrödinger equation for this system, in the stationary form:
\begin{equation}
    -\frac{\hbar
^2}{2m} \frac{d^2 \psi(x)}{dx^2} + V(x)\psi(x) = E\psi(x).\label{Schrödinger_equation}
\end{equation}

We assume that the potential energy is always measured relative to the particle's energy, which is effectively set to zero. We introduce 2000 sets of random potential functions \( V(x) \), keeping \( V(x) < 0 \) to ensure that the particles remain extended. For details on the random potential functions, see Appendix \ref{app:Appendix C} . By setting the normalization \({\hbar
^2}/{2m}=1\), Eq.~(\ref{Schrödinger_equation}) can be simplified for numerical convenience:
\begin{equation}
    \frac{d^2}{dx^2} \psi(x) = V(x) \psi(x).\label{Schrödinger equation}
\end{equation}
The density distribution \( \rho(x) \) is defined as:
\begin{equation}
    \rho(x) = |\psi(x)|^2.
\end{equation}

\begin{figure}[!t]
    \centering
    \captionsetup{font=footnotesize, labelfont=bf}

    \begin{minipage}[b]{0.64\textwidth}
        \centering
        \begin{minipage}[b]{0.49\textwidth}
            \centering
            \includegraphics[width=\textwidth]{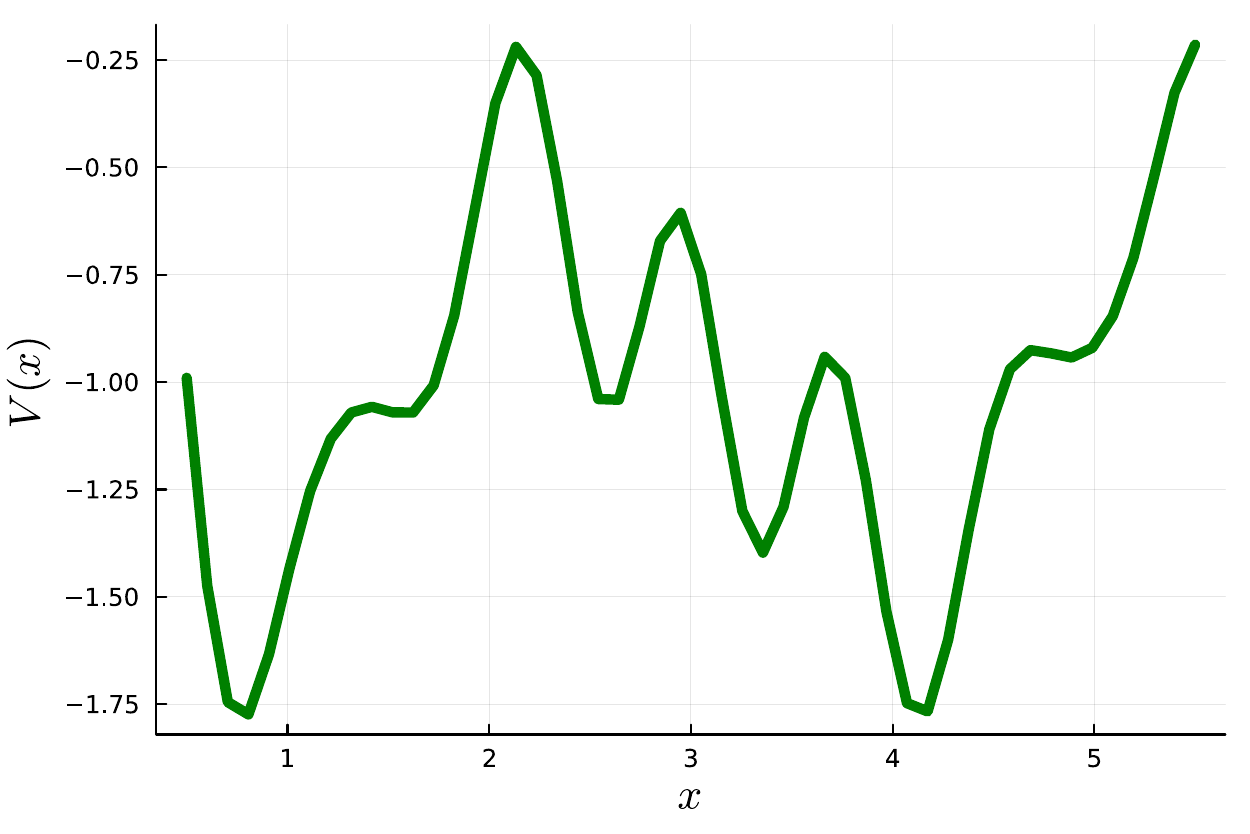}
        \end{minipage}
        \hfill
        \begin{minipage}[b]{0.49\textwidth}
            \centering
            \includegraphics[width=\textwidth]{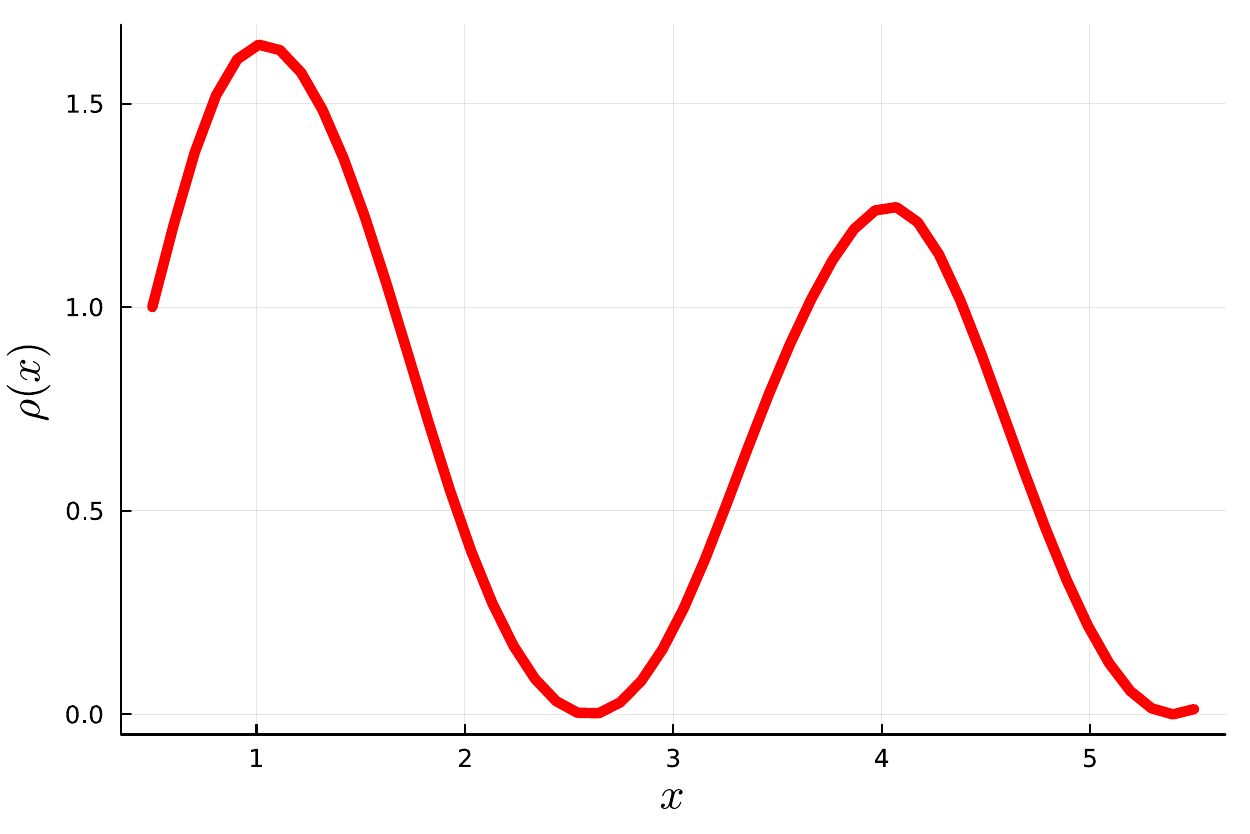}
        \end{minipage}
        \subcaption{}
    \end{minipage}
    \hfill
    \begin{minipage}[b]{0.32\textwidth}
        \centering
        \includegraphics[width=\textwidth]{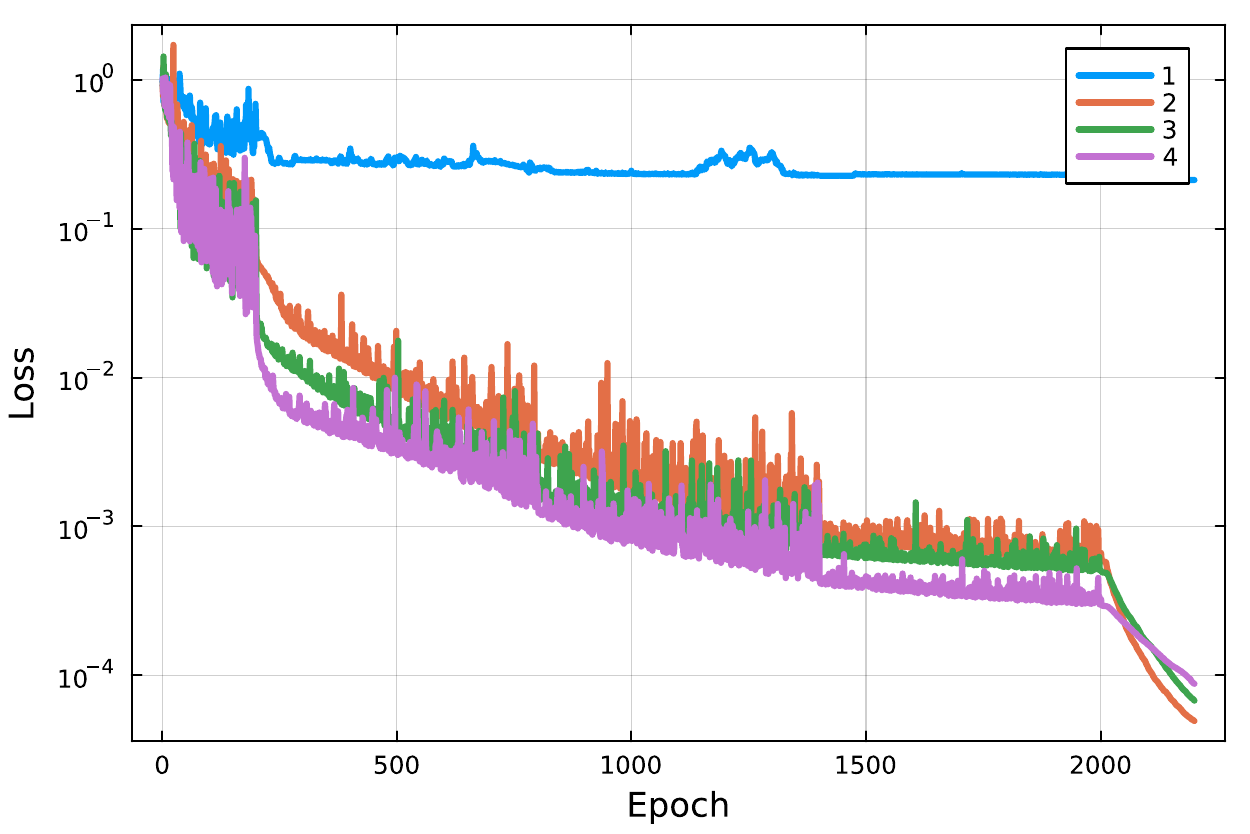}
        \subcaption{}
    \end{minipage}

    \begin{minipage}[b]{0.48\textwidth}
        \centering
        \includegraphics[width=\textwidth]{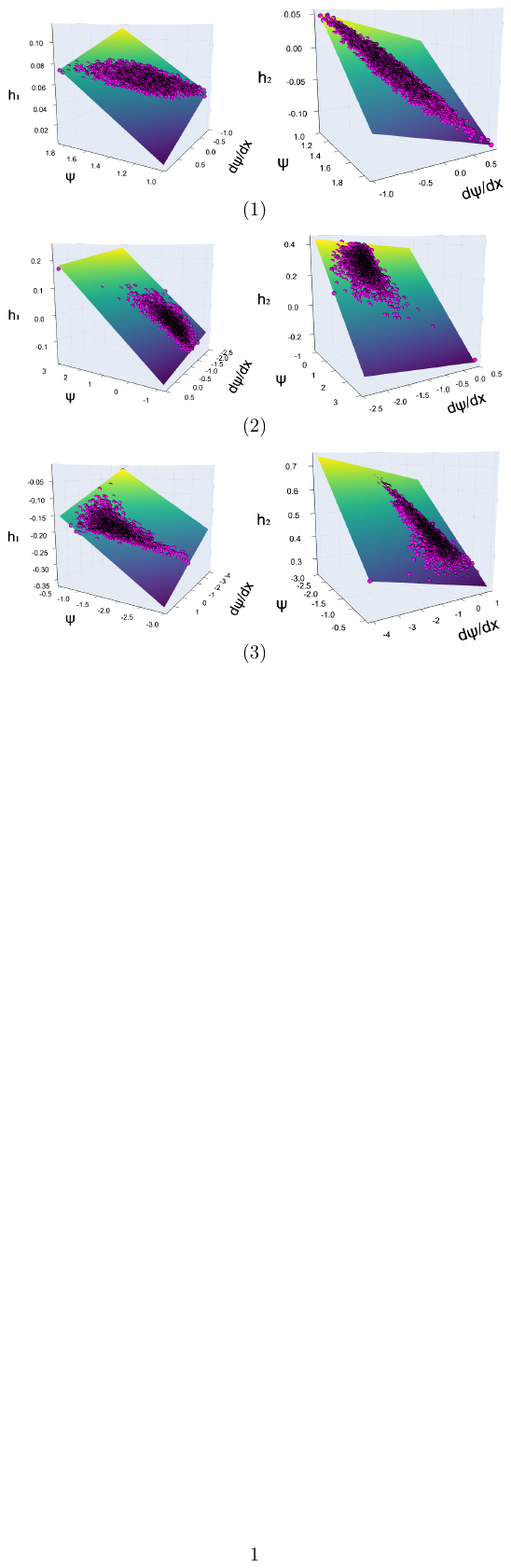}
        \subcaption{} 
    \end{minipage}
    \hfill
    \begin{minipage}[b]{0.5\textwidth}
        \centering
        \includegraphics[width=\textwidth]{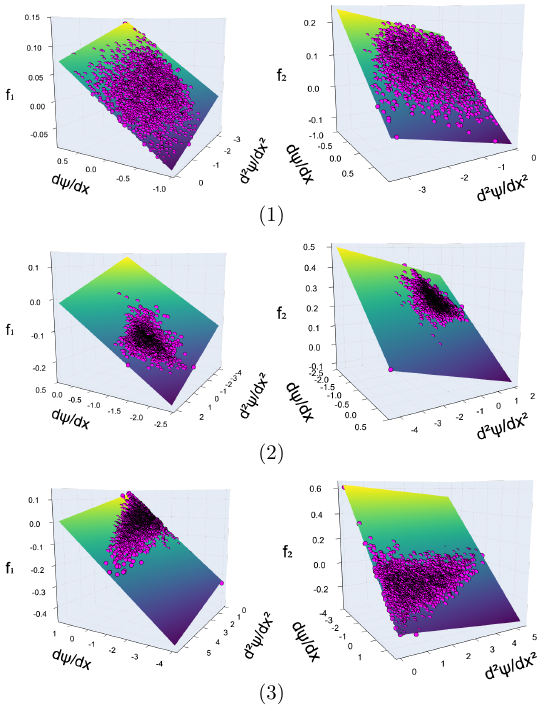}
        \subcaption{}
    \end{minipage}

    \caption{Wave function and Schrödinger equation.  
(a) Random potential and corresponding density distribution.  
(b) Impact of latent dimensions on loss, showing a two-dimensional space as optimal. 
(c) The latent representations store linear combinations of \( \psi(x) \) and \( {d\psi(x)}/{dx} \) at the 10th (1), 25th (2), and 40th (3) positions.    
(d) The governing functions learned by Neural ODEs are the linear combinations of  \( {d\psi(x)}/{dx} \) and \({d^2 \psi(x)}/{dx^2}\) at the 10th (1), 25th (2), and 40th (3) positions. The  \({d^2 \psi(x)}/{dx^2}\) denotes the right-hand side of Eq.~(\ref{Schrödinger equation}). In (c) and (d), the learned \(h\) and \(f\) appear as discrete points, which lie entirely on the reference plane.}
    \label{fig:Schrödinger Equation Model}
\end{figure}

We initialize both the wave function and its derivative with a value 1, evolve them over position \(x\), and calculate the corresponding probability density at each position to generate the training data. The input and label data are the trajectories of all 2000 sets of density distribution \( \rho(x) \) observed in the sequential 50 positions. Fig.~\ref{fig:Schrödinger Equation Model}(a) displays a random potential function \( V(x) \) and its corresponding density distribution \( \rho(x) \). The differential equations to be discovered are set as:
\begin{equation}
    \frac{d{h}(x)}{dx} = {f}({h}(x), V(x);\zeta)\label{29},
\end{equation}
where \(V(x)\) is a control variable.

Ablation experiments, as shown in Fig.~\ref{fig:Schrödinger Equation Model}(b), indicate that the optimal number of physical concepts  is two, a conclusion consistent with the results reported in \cite{wang2019emergent}. Fig.~\ref{fig:Schrödinger Equation Model}(c) shows that the learned representations at the 10th, 25th, and 40th positions store linear combinations of the actual wave function \( \psi(x) \) and its derivative \( {d\psi(x)}/{dx} \):
\begin{equation}
    h_j(x) = a_j\cdot \psi(x) + b_j\cdot \frac{d\psi(x)}{dx}+ c_j.
\end{equation}
Here, \( j \) takes the values 1 and 2. This suggests that the neural network has identified the linear combinations of the wave function and its derivative as the optimal concepts for describing this physical system. To see what Neural ODEs have learned, we calculate the derivatives of two latent representations with respect to the position \( x \):
\begin{equation}
   f_j(x)= \frac{dh_j(x)}{dx} =a_j\cdot \frac{d\psi(x)}{dx}+ b_j\cdot\frac{d^2 \psi(x)}{dx^2}= a_j\cdot \frac{d\psi(x)}{dx}+ b_j\cdot[V(x)\cdot \psi(x)].
\end{equation}
As shown in Fig.~\ref{fig:Schrödinger Equation Model}(d) for the learned governing functions at the 10th, 25th, and 40th positions, the Neural ODEs have successfully captured the correct governing functions.

One may notice that the latent-space visualizations differ among the three preceding examples. These differences arise from the fact that different governing equations combined with varying initial conditions lead to distinct geometric projections on the \(x\text{-}y\) plane (representing phase space trajectories in physical systems) at the fixed moment. 
For the heliocentric system, the governing equation for heliocentric angles features linear translation. To plot Fig.~\ref{fig:Heliocentric Model of the Solar System}(c), two initial heliocentric angles are sampled at 50 equally spaced values within fixed intervals, forming a \(50 \times 50\) 2D grid. After temporal evolution, the projection on the \(x\text{-}y\) plane remains a translated grid structure. 
For Newton's law of gravity, the radial distance \(r_0\) is uniformly sampled, while the initial velocity is fixed. This constraint reduces the system to a single degree of freedom. At any time \(t\), the radial position \(r\) and its time derivative \( {dr}/{dt} \) are uniquely determined by functions of \(r_0\):
\begin{equation}
 r = y_1(r_0, t), \quad \frac{dr(t)}{dt}  = y_2(r_0, t).
\end{equation}
As \(r_0\) varies continuously, these quantities trace a smooth parametric curve in the \(r\text{-} {dr}/{dt} \) plane. For Schrödinger's wave mechanics, all samples share identical initial wavefunctions and their spatial derivatives, but employ distinct random potential functions \(V(x)\) across samples. Consequently, each sample follows a unique governing equation, resulting in uncorrelated solution trajectories. After the position changes, the \(\psi\text{-} {d\psi}/{dx} \) plane exhibits scattered discrete points rather than continuous curves or grid-like structures.

 \subsection{Pauli's spin-magnetic formulation}

The Stern-Gerlach experiment in 1922 revealed the splitting of silver atoms trajectories in a non-uniform magnetic field, challenging classical physics and prompting Pauli in 1924 to propose the classically non-describable two-valuedness\cite{giulini2008electron}. In 1925, Uhlenbeck and Goudsmit introduced the notion of electron spin, which Pauli later integrated into the formulation of quantum mechanics, leading to the introduction of Pauli matrices, the two-component spin wave function, and its interaction with electromagnetic field \cite{pauli1927quantenmechanik}.

Now, let us consider the following scenario: suppose the Stern-Gerlach experiment had employed a uniform magnetic field and only a single stripe would have appeared on the screen. Such a setup would likely have masked the quantization of angular momentum. However, spin and the Pauli equation describe intrinsic properties and fundamental laws of particles, which remain present and operative, even if they are not directly observable in experimental data. Without the angular momentum quantization revealed by the Stern-Gerlach experiment, the discovery of spin might have been significantly delayed. Here we will investigate whether AI can still reconstruct the relevant concepts and equations in such scenarios, thereby shedding light on the hidden principle.

Consider a silver atom moving in a uniform and weak magnetic field along the $z$-direction. The stationary Pauli equation can be written as:
\begin{equation}
    \frac{d^2}{dx^2} \begin{pmatrix} \psi_{1}(x) \\ \psi_{2}(x) \end{pmatrix} = (V(x) \pm B) \begin{pmatrix} \psi_{1}(x) \\ \psi_{2}(x) \end{pmatrix}.\label{Pauli equation}
\end{equation}
Similar to the example of Schrödinger's wave mechanics, we have introduced the random potential function \( V(x) \) and made some normalization for simplicity. For detailed information about Eq.~(\ref{Pauli equation}), see Appendix \ref{app:Appendix D}.

\begin{figure}[!b]
    \centering
    \captionsetup{font=footnotesize, labelfont=bf}
    \begin{minipage}[b]{0.32\textwidth}
        \centering
        \includegraphics[width=\textwidth]{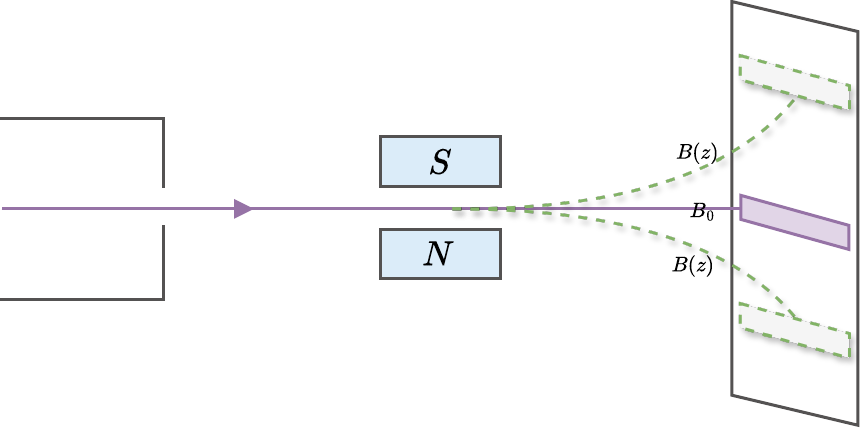}
        \subcaption{}
    \end{minipage}
    \hfill
    \begin{minipage}[b]{0.64\textwidth}
        \centering
        \begin{minipage}[b]{0.49\textwidth}
            \centering
            \includegraphics[width=\textwidth]{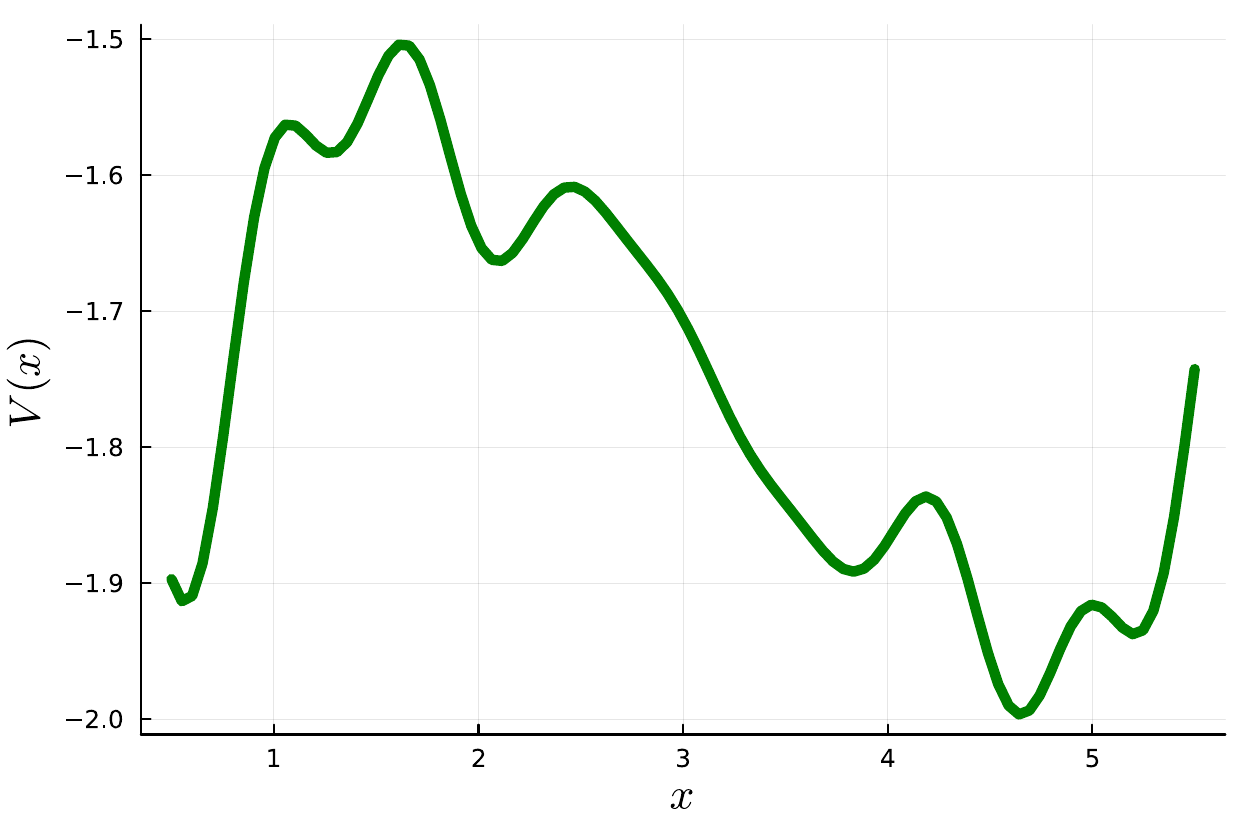}
        \end{minipage}
        \hfill
        \begin{minipage}[b]{0.49\textwidth}
            \centering
            \includegraphics[width=\textwidth]{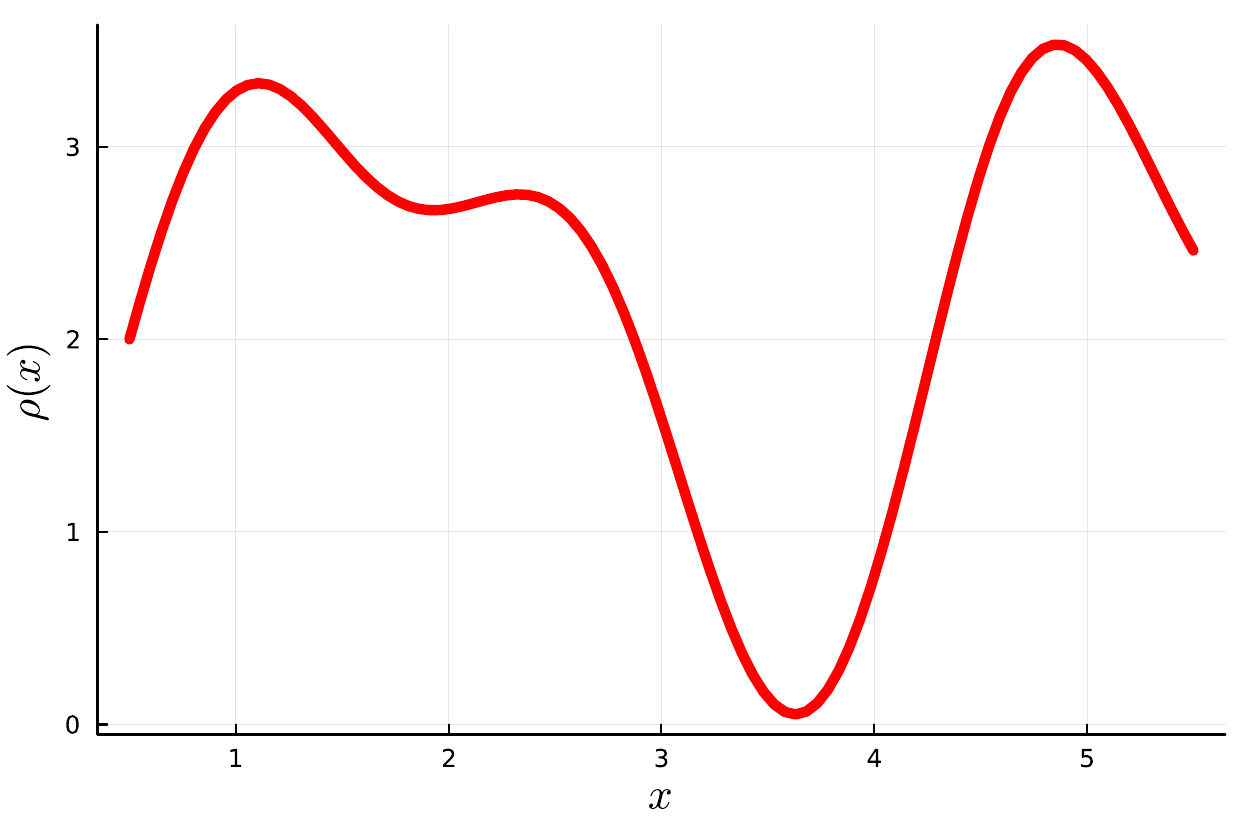}
        \end{minipage}
        \subcaption{}
    \end{minipage}
    
    \begin{minipage}[b]{0.32\textwidth}
        \centering
        \includegraphics[width=\textwidth]{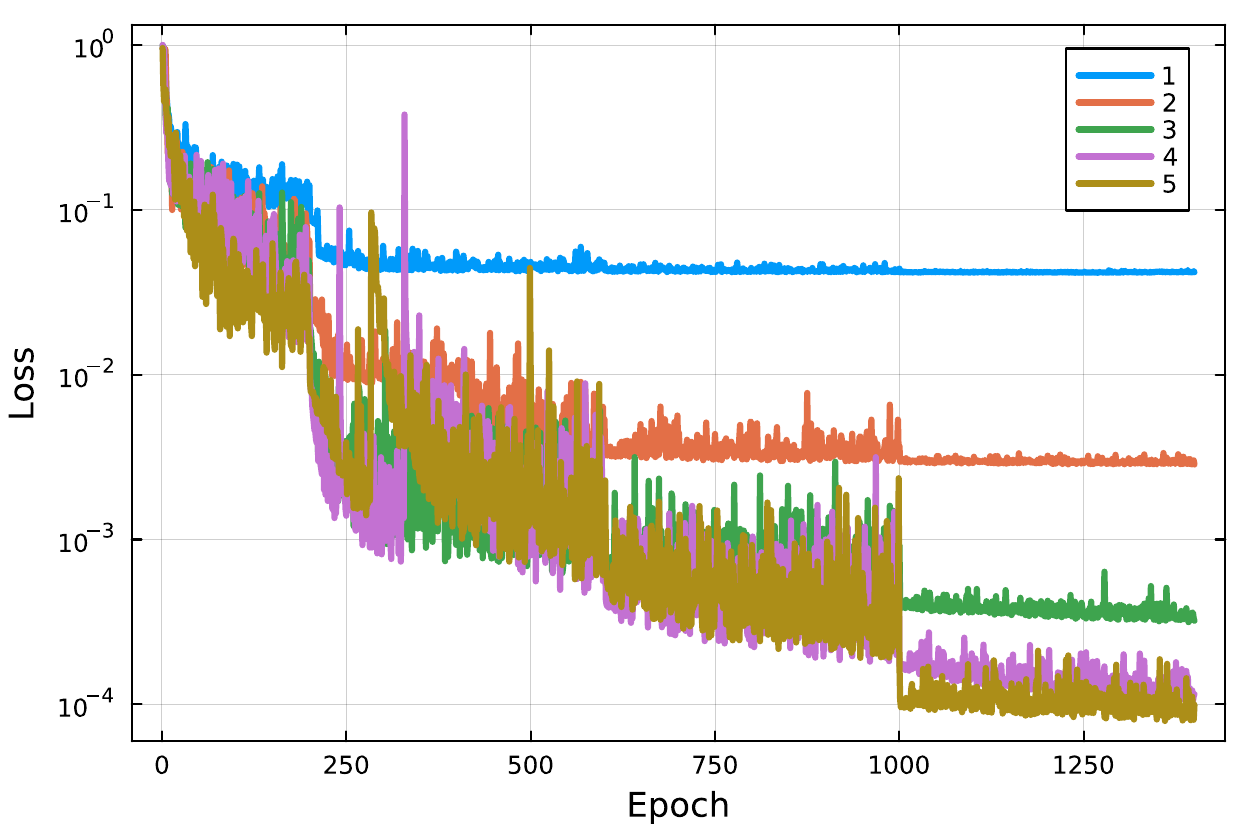}
        \subcaption{}
    \end{minipage}
    \hfill
    \begin{minipage}[b]{0.32\textwidth}
        \centering
        \includegraphics[width=\textwidth]{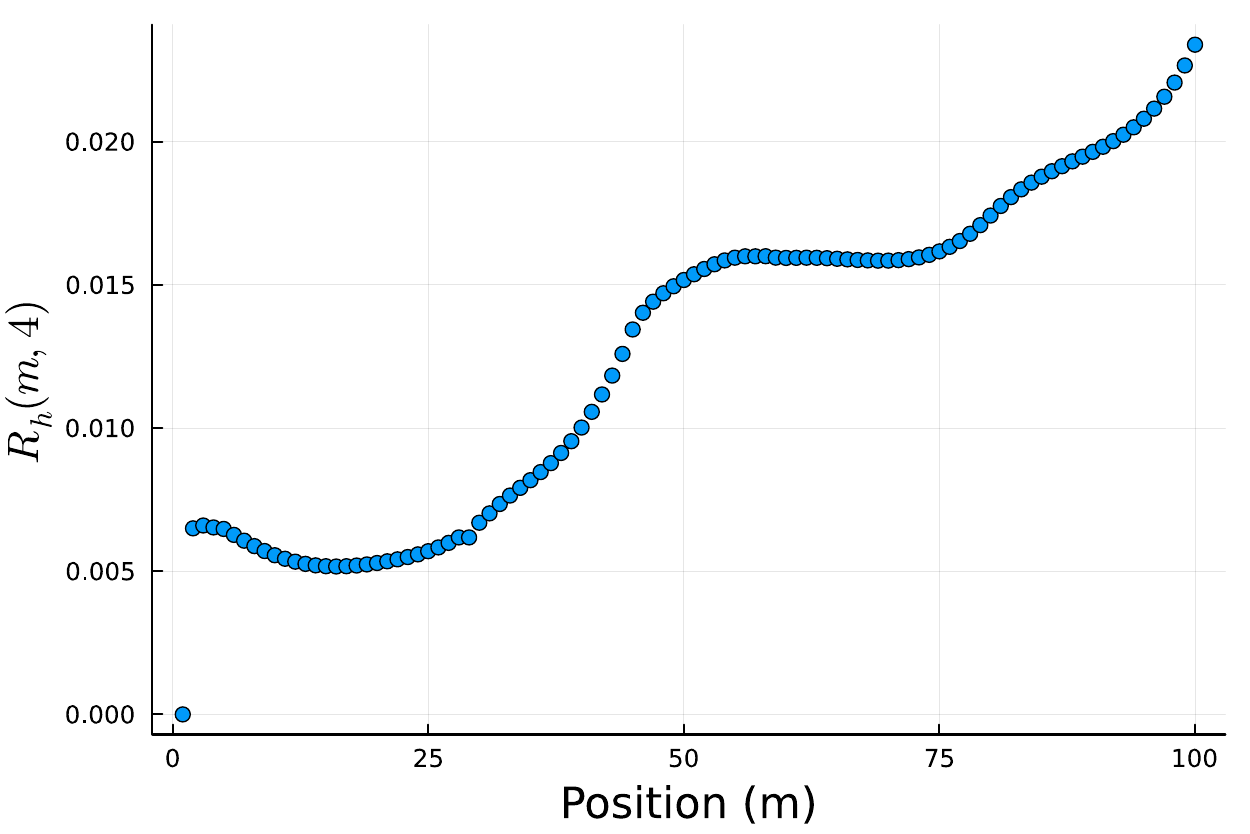}
        \subcaption{} 
    \end{minipage}
    \hfill
    \begin{minipage}[b]{0.32\textwidth}
        \centering
        \includegraphics[width=\textwidth]{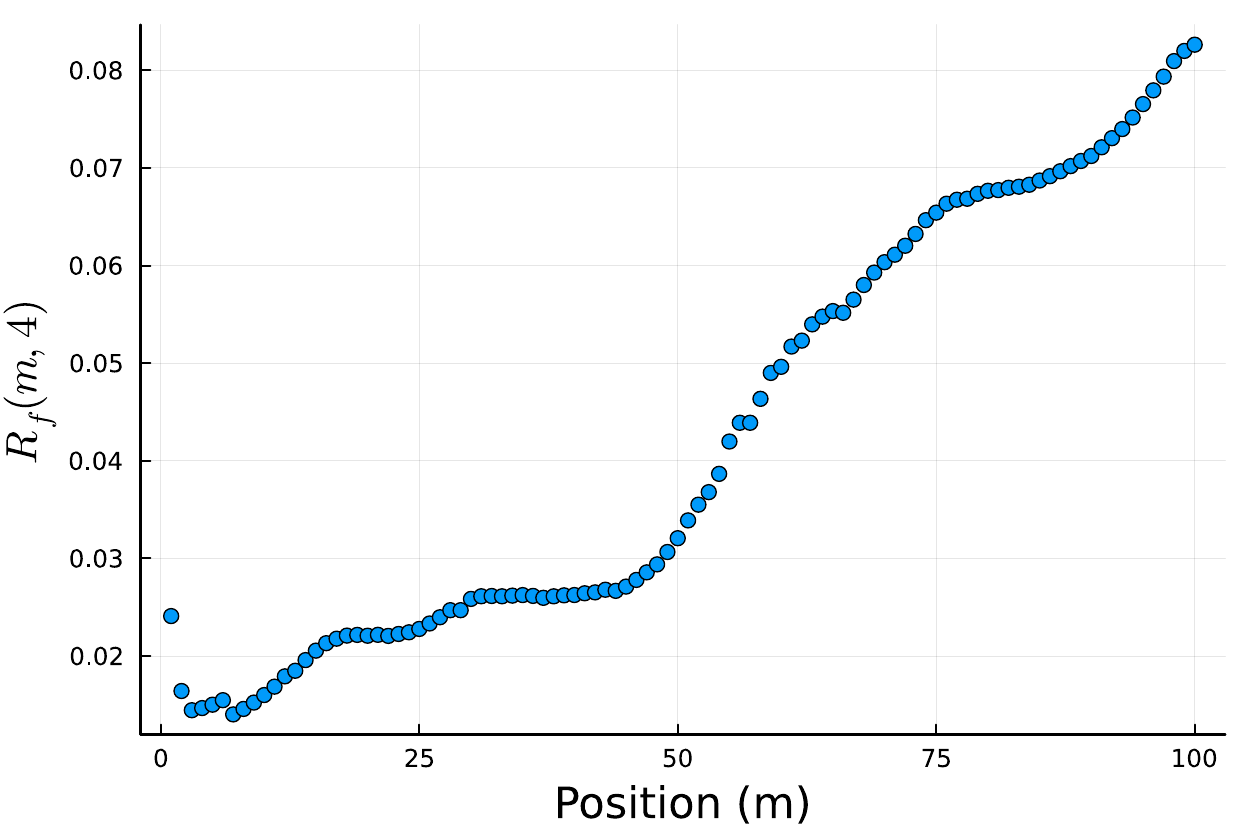}
        \subcaption{} 
    \end{minipage}
    
    \caption{Spin and Pauli equation.  
(a) Silver atoms in uniform (pink) and non-uniform (green) magnetic fields.
(b) Random potential and corresponding density distribution.  
(c) Impact of latent dimensions on loss, showing a four-dimensional space as optimal.  
(d) Relative error metric between the learned latent representations and the expected linear combinations of \( \psi_1(x) \) and \( \psi_2(x) \) as well as their derivatives. 
(e) Relative error metric between the learned governing functions and the expected linear combinations of \({d\psi_1(x)/dx}\), \({d\psi_2(x)/dx}\), \({d^2 \psi_1(x)}/{dx^2}\) and \({d^2 \psi_2(x)}/{dx^2}\), with the latter two referring to the right-hand side of Eq.~(\ref{Pauli equation}).
}
    \label{fig:Pauli Equation Model}
\end{figure}

We collect the probability density distribution on the screen as observed data. 
Since the magnetic field is uniform, each potential function \( V(x) \) corresponds to a unique density distribution, seen in Fig.~\ref{fig:Pauli Equation Model}(a) as a single stripe on the screen, which can be expressed as:
\begin{equation}
    \rho(x) = |\psi_{1}(x)|^2 + |\psi_{2}(x)|^2.\label{39}
\end{equation}

We initialize both the wave functions \( \psi_1(x) \) and \( \psi_2(x) \) and their derivatives to 1. Using the generated 4,000 sets of potential functions \( V(x) \) and setting \(B=1\) along with the Pauli equation, we produce 4,000 wave function trajectories, each consisting of 100 positions. By applying Eq.~(\ref{39}), we further compute the corresponding 4,000 probability density trajectories \( \rho(x) \), which serve as both the input and label for our neural network model. Fig.~\ref{fig:Pauli Equation Model}(b) displays a random potential function \( V(x) \) and its corresponding density distribution \( \rho(x) \).  The differential equations to be discovered are set as Eq.~(\ref{29}).

The results of the ablation experiment shown in Fig.~\ref{fig:Pauli Equation Model}(c) indicate that the optimal number of physical concepts required to describe this phenomenon is four. Similar to Schrödinger's wave mechanics, the network is expected to learn a linear combination of four concepts. Since multivariate linear combination systems are difficult to visualize directly, we define two relative error metrics to evaluate the performance of the model:
\begin{equation}
 R_{h}(m,n)=\frac{1}{mn}\sum_{i=0}^{m-1}{\sum_{j=1}^{n}} \frac{||h_{j}(x_{i})-\hat{h}_{j}(x_{i})||_{2}}{||h_{j}(x_{i})||_{2}},
    \end{equation}
 \begin{equation}
 R_{f}(m,n)=\frac{1}{mn}\sum_{i=0}^{m-1}{\sum_{j=1}^{n}} \frac{||f_{j}(x_{i})-\hat{f}_{j}(x_{i})||_{2}}{||f_{j}(x_{i})||_{2}}.
    \end{equation}
Here, \( m \) denotes the index of different  positions, and \(n \) indicates the dimension of the latent space. The latent representations are given by \({h}_{j}(x_i) = \left\{ {h}^{(k)}_{j}(x_i) \;\middle|\;k = 1, \dots, K \right\}\), where \( h^{(k)}_j(x_i) \) denotes the \(j\)-th latent representation at the \(i\)-th position of the \(k\)-th data sequence, while \( \hat{h}_j(x_i) \) denotes the predicted value obtained by fitting 4000 sample points. These predicted values are linear combinations of \( \psi_1(x) \), \( d\psi_1(x)/dx \), \( \psi_2(x) \), and \( d\psi_2(x)/dx \). Similarly, \( f_j(x_i) \) are the spatial derivatives of different latent representations, and \( \hat{f}_j(x_i) \) are their corresponding predicted values, which are linear combinations of \( d\psi_1(x)/dx \), \( d\psi_2(x)/dx \), \( d^2 \psi_1(x)/dx^2 \) and \( d^2 \psi_2(x)/dx^2 \). Note that the terms being summed are the \(L2\) norm of relative error \cite{horn2013matrix,hsu606002structural}.

The results presented in Fig.~\ref{fig:Pauli Equation Model}(d) and Fig.~\ref{fig:Pauli Equation Model}(e) indicate that, overall, the latent representations have successfully captured the linear combinations of the four concepts 
\begin{equation}
    h_j(x) = a_j \cdot \psi_1(x) + b_j\cdot \frac{d\psi_1(x)}{dx}+c_j \cdot \psi_2(x) + d_j\cdot \frac{d\psi_2(x)}{dx}+ e_j,
\end{equation}
and the Neural ODEs have identified the correct governing functions
\begin{align}
    f_j(x)=\frac{dh_j(x)}{dx} &= a_j \cdot \frac{d\psi_1(x)}{dx} + b_j \cdot \frac{d^2 \psi_1(x)}{dx^2} + c_j \cdot \frac{d\psi_2(x)}{dx} + d_j \cdot \frac{d^2 \psi_2(x)}{dx^2} \nonumber \\
    &= a_j \cdot \frac{d\psi_1(x)}{dx} + b_j \cdot \left[ (V(x) + B) \cdot \psi_1(x) \right] 
    \nonumber \\
     &+ c_j \cdot \frac{d\psi_2(x)}{dx} + d_j \cdot \left[ (V(x) - B) \cdot \psi_2(x) \right],
\end{align}
particularly at the front positions, as illustrated by \( R_f(50,4)\approx0.03 \).

For comparison, we have listed the two metrics for the four examples in Table \ref{tab:Relative error metrics}. Note that for the heliocentric system and the Newtonian gravity, \( m \) denotes the moment rather than position.
\begin{table}[ht]
\setlength{\tabcolsep}{10pt} 
\centering
\begin{tabular}{l c c c}
\toprule
Example & $(m, n)$& $R_{h}(m,n)$& $R_{f}(m,n)$\\
\midrule
Copernicus   & (50, 2)& 0.03 & 0.1\\
Newton       & (100, 2)& 0.01 & 0.03 \\
Schrödinger  & (50, 2)& 0.01 & 0.01 \\
Pauli        & (100, 4)& 0.02 & 0.08 \\
\bottomrule
\end{tabular}
\caption{Relative error metrics \(R_{h}(m,n)\) and \(R_{f}(m,n)\) for different physical systems.}
\label{tab:Relative error metrics}
\end{table}

\section{Conclusion and discussion }\label{app:conclusion and discussion}

 \subsection{Conclusion}
In this study, we extend the machine-learning model \textit{SciNet}\cite{iten2020discovering} , which can identify physical concepts by emulating human physical reasoning. Building upon \textit{SciNet}'s questioning mechanism, we integrate VAE and Neural ODEs to propose a neural network framework capable of simultaneously discovering physical concepts and governing equations. The applicability of this framework is demonstrated through four case studies inspired by the history of physics.

The former three cases include Copernicus’ heliocentrism, Newton’s law of gravity, and Schrödinger's wave mechanics, which have been previously studied in \cite{iten2020discovering,daniels2015automated,wang2019emergent} using different models. Now they have been re-addressed in the same framework. The fourth case is particularly notable, which involves the investigation of Pauli's spin-magnetic formulation. Our study highlights an intriguing difference from the historical development of physics, where the two splitting stripes observed in the Stern–Gerlach experiment under a non-uniform magnetic field clearly reveal distinct physical patterns. In contrast, our research shows that a machine can directly extract hidden physical patterns from experimental data obtained under a uniform magnetic field, where only a single stripe is observed, ultimately reconstructing essentially the same physical theory.

In summary, this study demonstrates the potential of machine learning to construct important physical theories directly from data, reducing reliance on background knowledge and lowering the requirement for the clarity of observed new physical patterns.

 \subsection{Discussion}

Our model faces several limitations in its application. Firstly, in the case of Pauli's spin-magnetic formulation, the performance of neural networks degrades as the space expands. This phenomenon may be attributed to the ``curse of length" problem in modeling underlying dynamics by gradient descent \cite{morrill2020neural,iakovlev2022latent}, where the complexity of the loss function increases as the observed trajectory becomes longer. Potential solutions to this problem could be explored in the future. For example, Iakovlev et al.~\cite{iakovlev2022latent} proposed using sparse Bayesian multiple shooting techniques integrated with Transformers within Neural ODEs models. This approach segments longer trajectories into shorter fragments and optimizes them in parallel, thereby improving efficiency and stability.

Secondly, although Neural ODEs are memory-efficient, the training method relying on numerical ODE solvers are computationally expensive, which confines current models to relatively small-scale systems.
Various strategies have been proposed to accelerate the training of Neural ODEs, some of which can be found in \cite{finlay2020train,daulbaev2020interpolation,ghosh2020steer,kelly2020learning,kidger2021hey,xia2021heavy,nguyen2022improving}. In particular, a novel training approach based on a variational formulation has recently been proposed, which performs global integrals without using the numerical ODE solvers \cite{zhaoaccelerating}. Additionally, Course and Nair~\cite{course2023state} proposed a reparameterization technique based on stochastic variational inference, enabling the identification of system states even when the governing equations are partially or completely unknown, thus closely aligning with the objectives of our work. Primarily, this method employs symbolic differential equations to represent the dynamics; however, it can also infer neural stochastic differential equations without invoking the numerical ODE solver when forward models do not require interpretability. It would be worthwhile to investigate in future research whether aspects of these methods could be effectively applied to the current model and the associated physics problems. 

Thirdly, our work is similar to the research in \cite{iten2020discovering}, which found that the latent representations store the linear combinations of real-world concepts. Ref.~\cite{nautrup2022operationally} explores a method for discovering independent concepts by proposing a neural network architecture where agents handling different aspects of a physical system can communicate relevant information efficiently. In Appendix \ref{app:Appendix E}, we attempt to address the same issue using a simple method. We found that, given the true equations can be organized as second-order equations\footnote {Note that the governing equations of most physical systems are second-order.}, our model successfully identifies independent concepts and corresponding equations for both the Newtonian gravity and the Schrödinger system. However, for the Pauli system, although the equation identification accuracy shows a slight improvement, the model does not capture the four independent concepts.

In the future, further improvements to the current model can also be explored.  For instance, we could design more effective questioning mechanisms to guide the learning process, impose multi-faceted constraints from physics and mathematics, and incorporate symbolic regression, particularly the recently proposed algorithm based on pretrained large language models \cite{shojaee2024llm}. Although these methods more or less introduce some prior knowledge, they would enhance both the interpretability of the framework and the accuracy of the discovered physical theories.

After sufficient improvements, we will follow the historical trajectory of physics and attempt to discover the Dirac equation and its associated spinor using machine learning. The Dirac equation, a significant successor to the Schrödinger and Pauli equations, is famously described by Wilczek as the most ``magical" equation in physics \cite{wilczek2004dirac}. Regardless of the outcome, we hope this exploration would provide profound insights for both AI and physics.

\section*{Acknowledgments}

We thank Sven Krippendorf, Tian-Ze Wu, Yu-Kun Yan, and Yu Tian for their valuable discussion. SFW was supported by NSFC grants (No.12275166 and No.12311540141).


\newpage \appendix*{Appendix} \setcounter{equation}{0} \renewcommand{%
\theequation}{A.\arabic{equation}}

\section{Network details and training process  \textbf{ } }\label{app:Appendix A}

\subsection{Network details}

Our model consists of three neural networks which need to be specified: the encoder, the governing functions in Neural ODEs, and the decoder. We set all of them as the fully connected neural networks. These neural networks used are specified by the observation input size, the number of latent neurons, output size, and the sizes of encoder, decoder, and the governing function in Neural ODEs, with their specific implementation details varying across application cases, as listed in Table \ref{Table 1}. In all cases, we use Kaiming initialization \cite{he2015delving} to initialize the network parameters. Our Neural ODEs are solved using the Tsitouras 5/4 Runge-Kutta method \cite{tsitouras2011runge}.
\begin{table}[ht]
    \setlength{\tabcolsep}{2pt}

    \begin{tabularx}{\textwidth}{C{0.25\linewidth}C{0.14\linewidth}C{0.12\linewidth}C{0.09\linewidth}C{0.18\linewidth}>{\centering\arraybackslash}X}
    \hline
    Example & Observation input & Latent neurons & Output & Encoder/ Decoder& \makecell{Governing\\functions}
\\
    \hline
    Copernicus            & 2    & 1-4    & 2 & (30, 30, tanh)  & (16, 16, tanh) \\
    Newton                & 1    & 1-4    & 1 & (64, 64, relu) & (16, 16, tanh) \\
    Newton (2nd Ord)      & 1    & 2    & 1 & (64, 64, relu)  & (16, 16, tanh) \\
    Schrödinger           & 50   & 1-4    & 1 & (64, 64, relu) & (16, 16, tanh) \\
    Schrödinger (2nd Ord) & 50   & 2    & 1 & (64, 64, relu)  & (16, 16, tanh) \\
    Pauli                 & 100  & 1-5    & 1 & (64, 64, relu)& (16, 16, tanh) \\
    \makecell{Pauli (2nd Ord)} & 100 & 4 & 1 & (64, 64, relu)& (16, 16, tanh) \\
    \hline
    \end{tabularx}
     \caption{Structure parameters of neural networks. The notation $(a,b,c)$ describes the first and second hidden layers and the activation function in the encoder, decoder, and the governing functions in Neural ODEs. The ``2nd Ord" indicates that the true equations are assumed to be second-order. Under this assumption, we attempt to capture independent physical concepts, as detailed in Appendix \ref{app:Appendix E}.} 
    \label{Table 1}
\end{table}

\subsection{Training process}

During the training process, neural network parameters are optimized using three optimizers RMSProp, Adam, and BFGS in sequences. For different application cases, the learning rates, the hyperparameter \(\beta\), and the total training epochs (including maximum iterations for BFGS) vary, as specified in Table \ref{tab:training-params}.
\begin{table}[ht]
\setlength{\tabcolsep}{2pt}
\centering

\begin{tabularx}{\textwidth}{C{0.25\linewidth}C{0.18\linewidth}C{0.18\linewidth}C{0.18\linewidth}C{0.18\linewidth}}
\toprule
Example & Batch size& Learning rate & \(\beta\) & Epoch \\ \midrule
Copernicus &64 & 0.01-0.0001 & 0.01 & 3000 \\
Newton  &64 & 0.01-0.001 & 0 & 2200 \\
Newton (2nd Ord) &64 & 0.01-0.001 & 0 & 2200 \\
Schrödinger &64 & 0.01-0.001 & 0.001 & 1600 \\
Schrödinger (2nd Ord) &64 & 0.01-0.001 & 0.001 & 1600 \\
Pauli &64 & 0.01-0.001 & 0.0001 & 2000 \\
Pauli (2nd Ord) &64 & 0.01-0.001 & 0.1 & 1400 \\ \bottomrule
\end{tabularx}
\caption{Parameters specifying the training process.}

\label{tab:training-params}
\end{table}

\section{VAE and disentangled latent representations
\textbf{ } }\label{app:Appendix B}
VAE \cite{kingma2013auto} extends conventional autoencoders by introducing a probabilistic framework: rather than directly encoding the input \({x}\) into a deterministic latent representation, VAE learns a distribution \(p_{{\phi}}({h}|{x})\), commonly assumed to be \(\mathcal{N}({\mu}, {\sigma})\). The corresponding decoder \(p_{{\theta}}({x}|{h})\) then reconstructs the original data by sampling \({h}\) from the latent space. To keep this latent distribution close to a selected prior, VAE introduces a KL divergence term \(D_{\mathrm{KL}}\bigl[p_{{\phi}}({h}|{x}) \,\big\|\, p({h})\bigr]\) as a regularizer.

VAE is trained by maximizing the evidence lower bound, a lower bound on the log-likelihood \(\log p_{{\theta}}({x})\) \cite{kingma2013auto}:
\[
\log p_{{\theta}}({x}) 
\;\ge\; 
L_{\mathrm{VAE}}({\theta},{\phi}; {x}) 
\;=\; 
\mathbb{E}_{{h}\sim p_{{\phi}}({h}|{x})}\bigl[\log p_{{\theta}}({x}|{h})\bigr] 
\;-\; 
D_{\mathrm{KL}}\bigl[p_{{\phi}}({h}|{x}) \,\big\|\, p({h})\bigr]\tag{B.1},
\]
where the first term can be interpreted as the negative reconstruction error,  and the second term is the negative KL divergence. Maximizing \(L_{\mathrm{VAE}}\) maintains reconstruction fidelity while constraining the latent representation to match the chosen prior distribution.

Building upon the standard VAE, \(\beta\)-VAE introduces a hyperparameter \(\beta \geq 0\) to scale the KL divergence term, thereby controlling the balance between reconstruction fidelity and latent-variable disentanglement \cite{higgins2017beta}. The optimization objective is given by
\[
L_{\beta\text{-VAE}}({\theta}, {\phi}; x, \beta) 
= \; 
-\mathbb{E}_{{h}\sim p_{{\phi}}(h|{x})}\bigl[\log p_{{\theta}}({x}|{h})\bigr]
\;+\beta \cdot \; 
D_{\mathrm{KL}}\bigl[p_{{\phi}}({h}|{x}) \,\big\|\, p({h})\bigr]\tag{B.2}.
\]
Here, the KL divergence encourages the posterior distribution \(p_{{\phi}}({h}|{x})\) to closely match a chosen prior distribution \(p({h})\), typically an isotropic Gaussian. This alignment promotes statistical independence among latent variables by penalizing deviations from a factorized prior distribution. Increasing \(\beta\) enhances this regularization, strongly discouraging correlations between latent factors and leading the model to represent data through distinct, non-redundant latent dimensions. Consequently, a higher \(\beta\) fosters more disentangled representations, though potentially at the expense of reduced reconstruction accuracy.

Under the assumptions that the posterior distribution satisfies \(p_{{\phi}}({h}|{x}) = \mathcal{N}(\mu_j, \sigma_j )\), the prior distribution satisfies \(p({h}) = \mathcal{N}({0},{\sigma_h
})\)\footnote {In our numerical implementation, we set the prior standard deviation as $\sigma_h = 0.1$.}, and the decoder outputs a multivariate Gaussian with mean \( {\hat{x}}(t)\) and fixed covariance matrix \(\hat{\sigma} = \frac{1}{\sqrt{2}}\,\mathbbm{1}\), together with the Neural ODEs parameters \(\zeta\), the final loss can be written as
\[
L(x; {\phi}, {\zeta}, {\theta}) 
= 
 \frac{1}{I J}\sum_{j=1}^{J}\sum_{i=0}^{I-1}\left\|{x}_{j}(t_i)-\hat{x}_{j}(t_i)\right\|^{2}_2+\;
  \frac{\beta}{2} \sum_{j}
    \bigl(\frac{\mu_j^{2}}{\sigma_h^2} + \frac{\sigma_j^{2}}{\sigma_h^2} - \log \sigma_j^{2})
\tag{B.3},
\]
up to an irrelevant constant.

In our model, the VAE does not inherently guarantee that each latent dimension corresponds directly to standard physical concepts. Instead, it may result in latent dimensions capturing linear combinations of the physical concepts. To further guide the neural network toward identifying explicitly independent concepts, additional constraints can be introduced, see Appendix \ref{app:Appendix E} for an attempt.

\section{Random potential\textbf{ } }\label{app:Appendix C}
We generate the random potential function \(V(x)\) for the Schrödinger equation as follows:
 \begin{enumerate}

    \item Generate a set of random coefficients:
   \begin{equation}
   v_{\text{coef}} = [c_1, \ldots, c_k, \ldots, c_9] \tag{C.1},
   \end{equation}
   where \(c_k\) is drawn from a standard normal distribution.
    \item Construct the random potential by expanding it in a series of sinusoidal basis functions:
   \begin{equation}
   V(x) = \sum_{k=1}^9 c_k \sin(kx) - 1 \tag{C.2},
   \end{equation}
where the constant term \(-1\) is introduced merely to shift the random potential overall so that it is more likely to satisfy the following condition. 
    \item We select the potential function by requiring:
   \begin{equation}
   -3 < V(x) < 0\tag{C.3}\label{c.3}.
   \end{equation}
If Eq. (\ref{c.3}) holds, we collect the corresponding coefficients into a specified set. This process continues until  \(K\) sets of desired coefficients are collected. 
 
\end{enumerate}

In the Pauli system, we choose the range of the random potential function to be [-2, -1.5], with a magnetic field \(B=1\). While these parameters are not unique, their selection considered the following: First, the range should be limited, as a large range would excessively increase the wave function's oscillation frequency, complicating the numerical solving and the training of Neural ODEs. Second, the range should not be too narrow, as insufficient variation would hinder the model's ability to learn the system's diverse characteristics.

\section{Pauli equation for silver atoms in a uniform weak magnetic field\textbf{ } }\label{app:Appendix D}

In this section, we begin with introducing the Pauli equation in a general form and then turn to the case of the silver atoms in a uniform weak magnetic field.

Consider a particle with mass \( m \) and charge \( q \) in an electromagnetic field with vector potential  \( \mathbf{A} \)  and scalar potential $\phi$. The Pauli equation has a general form\cite{landau1977quantum}:
\[
i \hbar \frac{\partial}{\partial t} \Psi( \mathbf{r},t) = \left[ \frac{1}{2m} \left[\bm{\sigma} \cdot (\hat{\mathbf{p}} - q \mathbf{A})\right]^2 + q\phi \right] \Psi(\mathbf{r},t) \tag{D.1}\label{eqD1}.
\]
Here, \( \bm{\sigma} = (\sigma_x, \sigma_y, \sigma_z) \) are the Pauli matrices, \( \hat{\mathbf{p}} = -i\hbar \nabla \) is the momentum operator and the Pauli spinor $\Psi(\mathbf{r},t) = \begin{bmatrix} \Psi_1(\mathbf{r},t) \\ \Psi_2(\mathbf{r},t) \end{bmatrix}$ has two components, representing the spin-up state and the spin-down state.

If the magnetic field is weak, the Pauli equation (\ref{eqD1}) can be rewritten as:
\[
i \hbar \frac{\partial}{\partial t} \begin{bmatrix} \Psi_1(\mathbf{r},t) \\ \Psi_2(\mathbf{r},t) \end{bmatrix} =\frac{1}{2m} \left[\hat{\mathbf{p}}^2 - q( \hat{\mathbf{L}} + 2\mathbf{S}) \cdot \mathbf{B}+ q\phi \right]\begin{bmatrix} \Psi_1(\mathbf{r},t) \\ \Psi_2(\mathbf{r},t) \end{bmatrix} \tag{D.2}\label{eqD2},
\]
where $\hat{\mathbf{L}}$ is the orbital angular momentum and \( \mathbf{S} =  \hbar\bm{\sigma}/2\) is the spin. 

Now we focus on an electrically neutral silver atom, which has one external electron in the state with orbital angular momentum $\hat{\mathbf{L}}=0$ and the intrinsic magnetic moment \( \mu_B = {e\hbar}/{2m} \). Then Eq.~(\ref{eqD2}) can be reduced to\cite{basdevant2007lectures}:
\[
i \hbar \frac{\partial}{\partial t} \begin{bmatrix} \Psi_1(\mathbf{r},t) \\ \Psi_2(\mathbf{r},t) \end{bmatrix} = \left(\frac{\hat{\mathbf{p}}^2}{2m} - \mu_B \, \mathbf{B}\cdot\bm{\sigma}\right)\begin{bmatrix} \Psi_1(\mathbf{r},t) \\ \Psi_2(\mathbf{r},t) \end{bmatrix}. \tag{D.3}\label{eqD3}
\]

Similar to the example of the Schrödinger system, we introduce the random potential function \( V(\mathbf{r}) \) as follows: 
\[
i \hbar \frac{\partial}{\partial t} \begin{bmatrix} \Psi_1(\mathbf{r},t) \\ \Psi_2(\mathbf{r},t) \end{bmatrix} =\left(\frac{\hat{\mathbf{p}}^2}{2m}  - \mu_B \, \mathbf{B}\cdot\bm{\sigma} + V(\mathbf{r})\right)\begin{bmatrix} \Psi_1(\mathbf{r},t) \\ \Psi_2(\mathbf{r},t) \end{bmatrix} \tag{D.4}\label{eqD4}.
\]
Furthermore, we suppose $\mathbf{B}$ is a uniform magnetic field along the $z$-direction (\( \mathbf{B} = B_z \hat{k} \)) and involve a potential only with \( x \)-dependence. After separating variables in the wave function:
\[
\begin{bmatrix} \Psi_1(x,t) \\ \Psi_2(x,t) \end{bmatrix} = \begin{bmatrix} \psi_1(x) \\ \psi_2(x) \end{bmatrix} \phi(t) \tag{D.5}\label{eqD5},
\]
we obtain the stationary equations:
\begin{equation}
-\frac{\hbar^2}{2m} \frac{d^2}{dx^2} \psi_1(x) = \left(E +  \mu_BB_z - V(x)\right) \psi_1(x) \tag{D.6}\label{eqD6},
\end{equation}
\begin{equation}
-\frac{\hbar^2}{2m} \frac{d^2}{dx^2} \psi_2(x) = (E - \mu_BB_z - V(x)) \psi_2(x) \tag{D.7}\label{eqD7}.
\end{equation}

For convenience in machine learning, we set \( E  = 0 \), \( \hbar^2/2m = 1 \) and \( \mu_BB_z  = B \), which lead to our target Pauli equations: 
\begin{equation}
\frac{d^2}{dx^2} \psi_1(x) = (V(x) + B) \psi_1(x) \tag{D.8}\label{eqD8},
\end{equation}
\begin{equation}
\frac{d^2}{dx^2} \psi_2(x) = (V(x) - B) \psi_2(x) \tag{D.9}\label{eqD9}.
\end{equation}
Without loss of generality, we set \(B = 1\).

\section{Capturing independent physical concepts \textbf{ } }\label{app:Appendix E}

In the main text, we demonstrated that the neural network can recognize linear combinations of physical concepts. Here we try to enable the network to discover independent physical concepts. To achieve this, we introduce a universal constraint that the true governing equations in the last three cases are second-order.

\subsection{Newton's law of gravity}
The ablation experiment informs us that the dimension of the latent space is 2. Then we assume the system is governed by a second-order equation, which can be equivalently formulized as:
\begin{equation}
    \frac{dh_1(t)}{dt} = h_2(t), \quad \frac{dh_2(t)}{dt} =  {f}( h_1(t), h_2(t), r_0;\zeta)\tag{E.1}.
\end{equation}
Here \( h_1 \) and \( h_2 \) are the two latent states.

\begin{figure}[ht]
    \centering
    \captionsetup{font=footnotesize, labelfont=bf}
    \setlength{\abovecaptionskip}{0pt}
    \setlength{\belowcaptionskip}{0pt}

    \begin{minipage}[b]{0.64\textwidth}
        \centering
        \includegraphics[width=\textwidth]{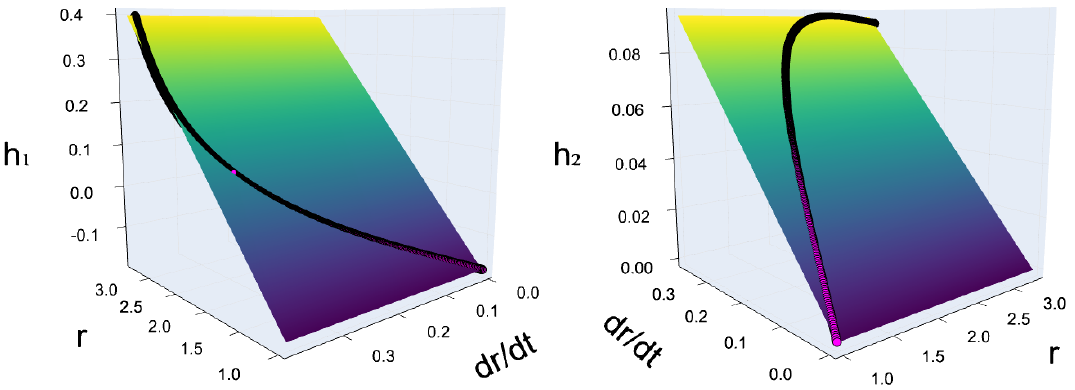}
        \subcaption*{(1)}
        \vspace{5pt}

        \includegraphics[width=\textwidth]{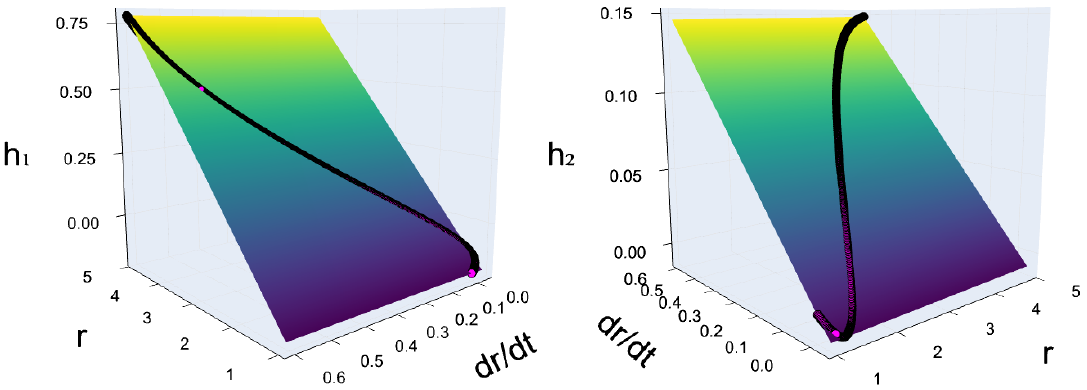}
        \subcaption*{(2)}
        \vspace{5pt}

        \includegraphics[width=\textwidth]{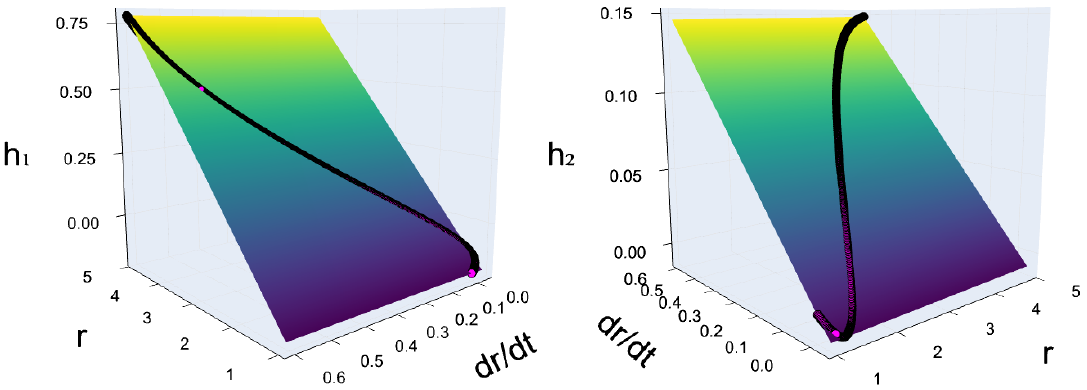}
        \subcaption*{(3)}
        \subcaption{}
    \end{minipage}\hfill
    \begin{minipage}[b]{0.35\textwidth}
        \centering
        \includegraphics[width=\textwidth]{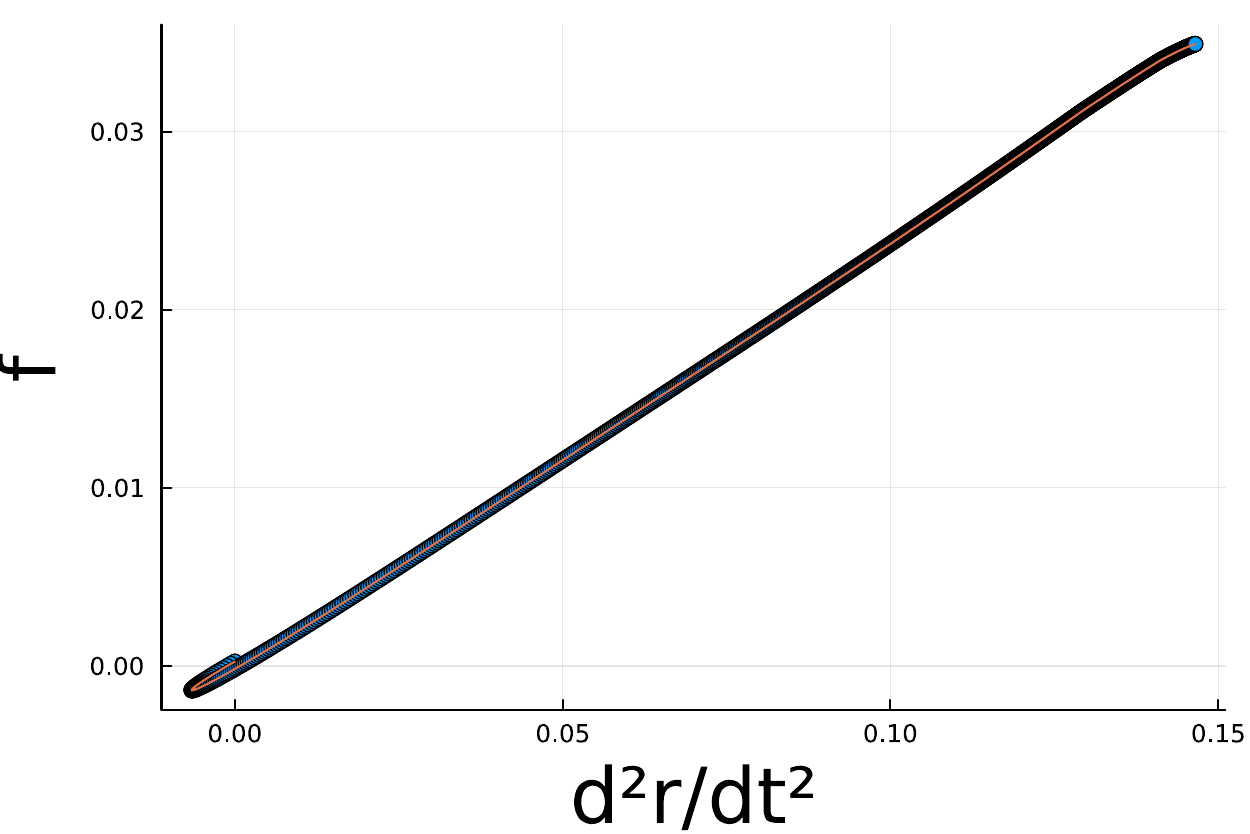}
        \subcaption*{(1)}
        \vspace{5pt}

        \includegraphics[width=\textwidth]{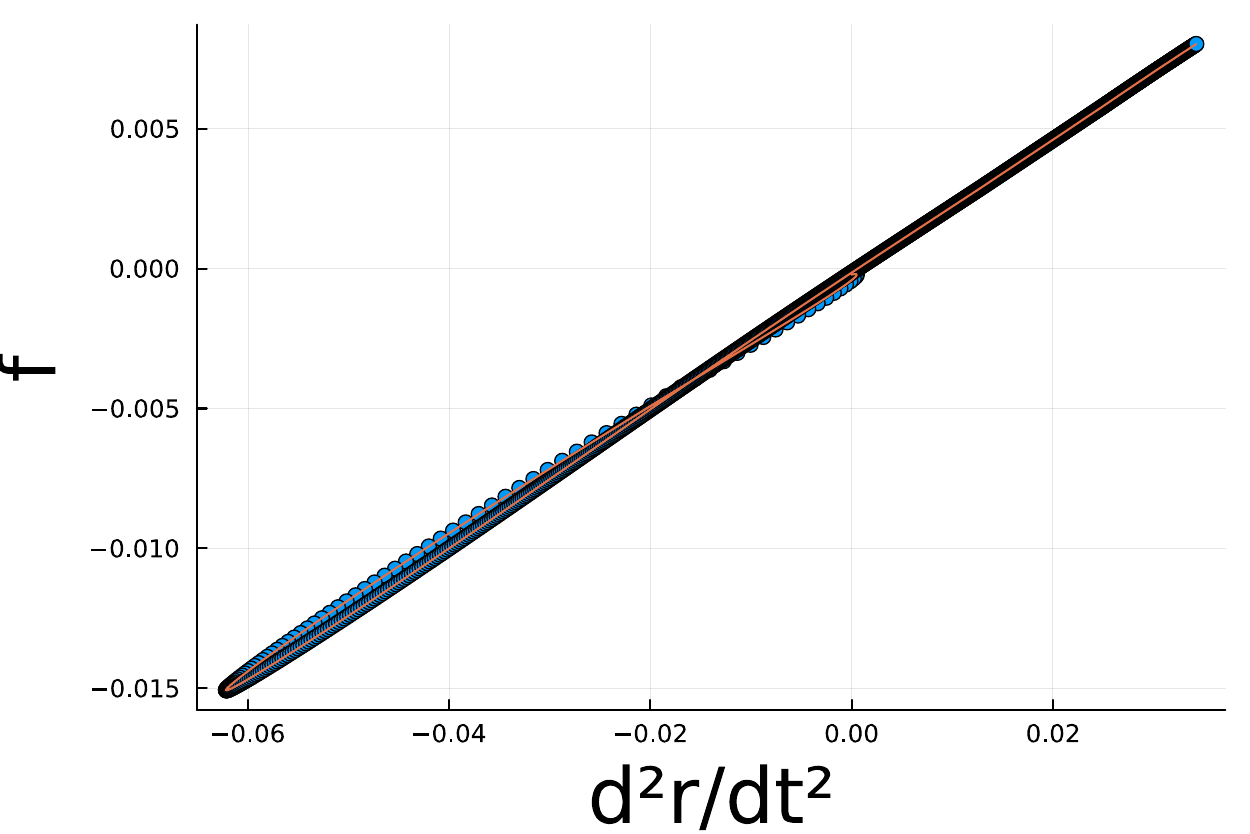}
        \subcaption*{(2)}
        \vspace{5pt}

        \includegraphics[width=\textwidth]{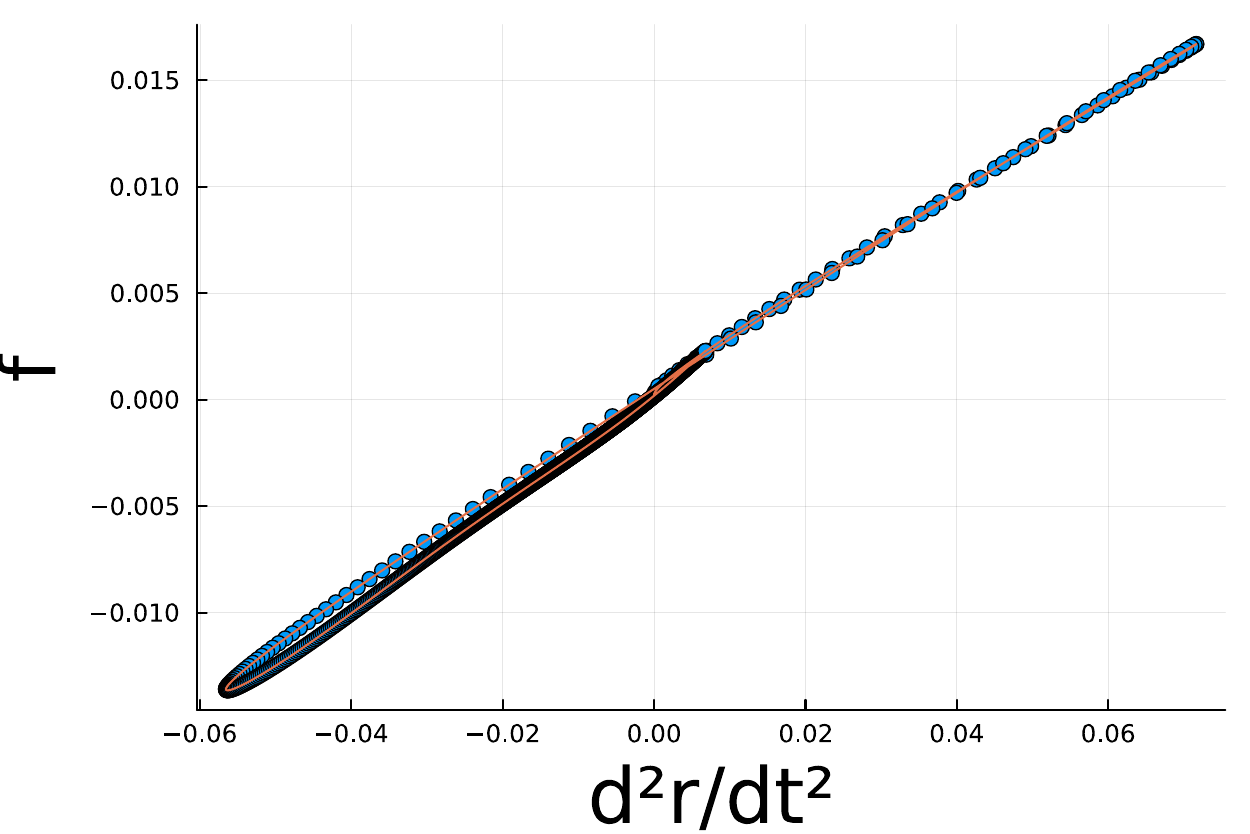}
        \subcaption*{(3)}
        \subcaption{}
    \end{minipage}

    \caption{Law of gravity with the second-order assumption. (a) Two learned latent representations at the 20th (1), 50th (2), and 80th (3) moments. (b) The governing functions learned by Neural ODEs at the 20th (1), 50th (2), and 80th (3) moments.}
    \label{fig:Model of Universal Gravitation independent}
\end{figure}

We exhibit the training results with this setting as follows. Fig.~\ref{fig:Model of Universal Gravitation independent}(a) shows that at the 20th, 50th, and 80th moments, the two latent representations no longer store linear combinations of physical concepts, but instead discover the two independent concepts \( r \) and \( dr/dt \)\footnote {This can be observed from the bottom edges of the planes in  Fig.~\ref{fig:Model of Universal Gravitation independent}(a), which are parallel to either \( r \) or \( dr/dt \).}. They differ from the true physical quantities only by a linear transformation:
\begin{equation}
   h_1(t) = a_1\cdot r(t) + b_1, \quad  h_2(t) = a_2\cdot \frac{dr(t)}{dt} + b_2\tag{E.2}.
\end{equation}
Accordingly, the equation learned by Neural ODEs is expected to be:
\begin{equation}
     {f} = \frac{d h_2(t)}{dt} = \frac{d}{dt} \left[a_2 \cdot \frac{dr(t)}{dt} + b_2 \right] = a_2 \cdot \frac{d^2 r(t)}{dt^2} = a_2\cdot \left[\frac{1}{r^2} \cdot\left(\frac{r_0^2}{r} - 1\right)\right]\tag{E.3}.
\end{equation}
As shown in Fig.~\ref{fig:Model of Universal Gravitation independent}(b), there is actually a linear relationship between the governing functions $f$ learned by Neural ODEs at the 20th, 50th, and 80th moments and the expected functions \( {d^2 r}/{dt^2} \).

\subsection{Schrödinger's wave mechanics}
Similar to the previous example,  we assume that the latent differential equation is second-order. Then we set:
\begin{equation}
    \frac{dh_1(x)}{dx} = h_2(x), \quad \frac{dh_2(x)}{dx} = {f}(h_1(x), h_2(x), V(x);\zeta)\tag{E.4}.
\end{equation}
Here \( h_1 \) and \( h_2 \) are two latent states.

\begin{figure}[!ht]
    \centering
    \captionsetup{font=footnotesize, labelfont=bf}

    \begin{minipage}[b]{0.64\textwidth}
        \centering
        \includegraphics[width=\textwidth]{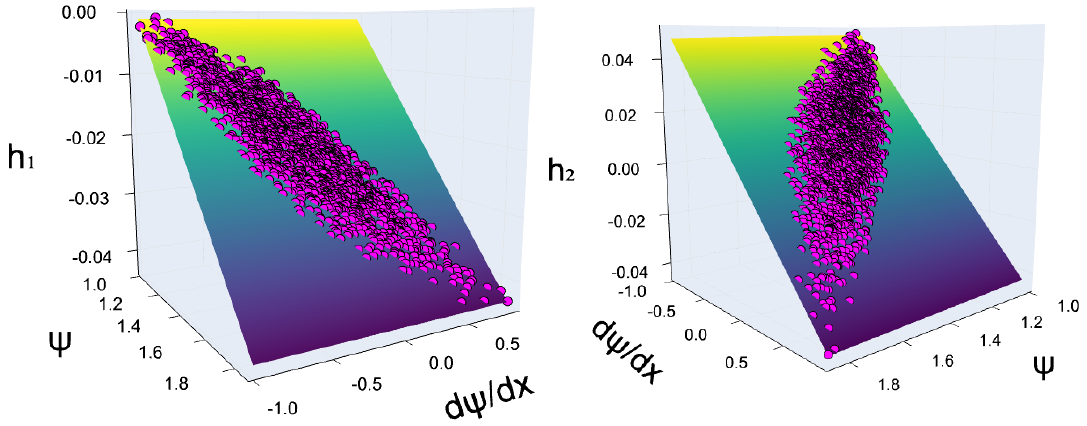}
        \subcaption*{(1)}
        \vspace{5pt}

        \includegraphics[width=\textwidth]{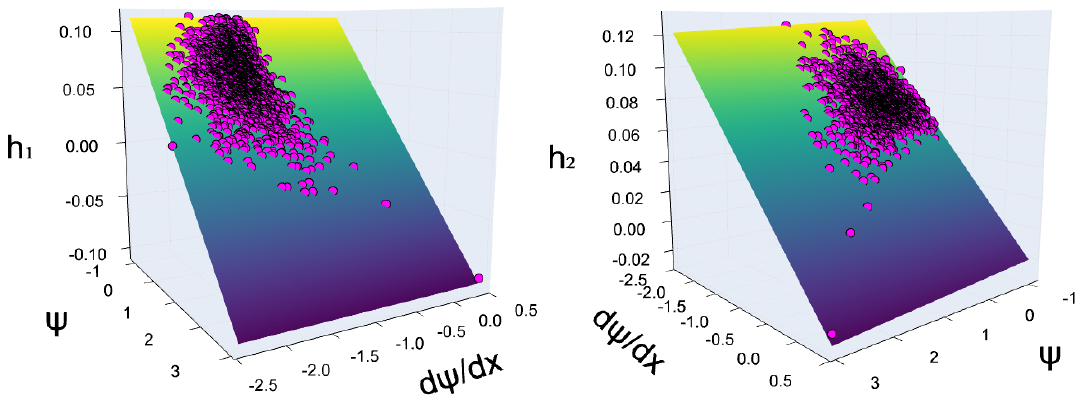}
        \subcaption*{(2)}
        \vspace{5pt}

        \includegraphics[width=\textwidth]{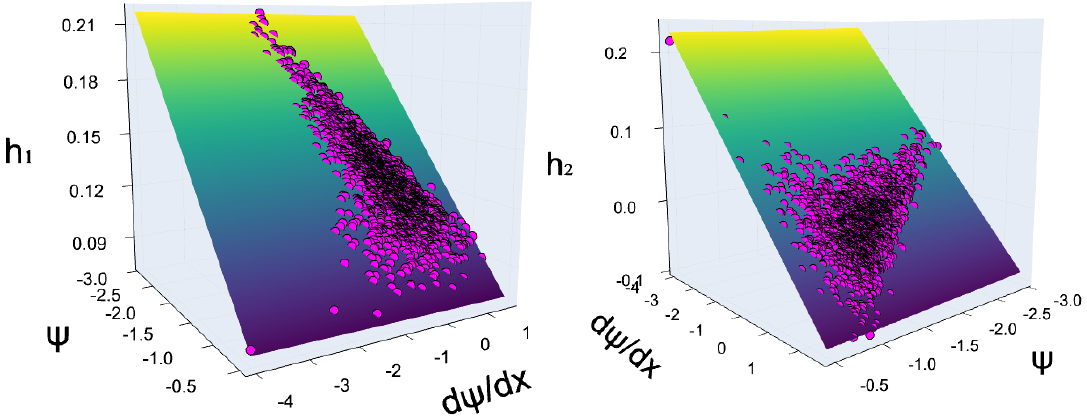}
        \subcaption*{(3)}
        \subcaption{}
    \end{minipage}\hfill
    \begin{minipage}[b]{0.33\textwidth}
        \centering
        \includegraphics[width=\textwidth]{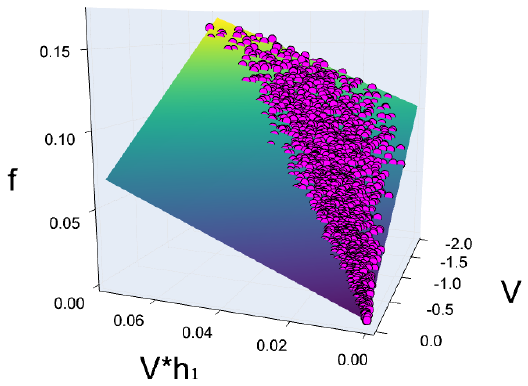}
        \subcaption*{(1)}
        \vspace{5pt}

        \includegraphics[width=\textwidth]{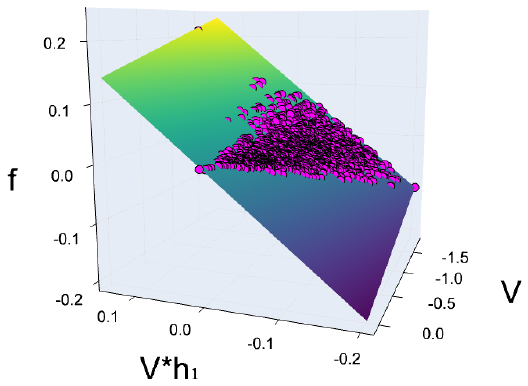}
        \subcaption*{(2)}
        \vspace{5pt}

        \includegraphics[width=\textwidth]{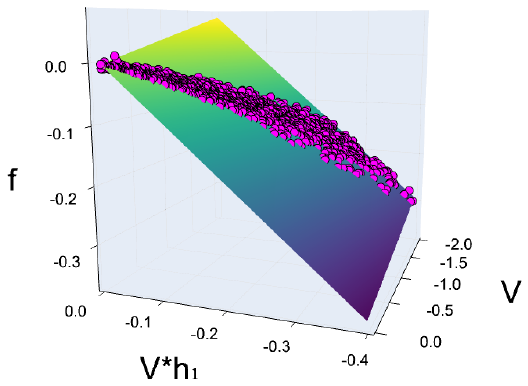}
        \subcaption*{(3)}
        \subcaption{}
    \end{minipage}

    \caption{The Schrödinger system with the second-order assumption. (a) Two learned latent representations at the 10th (1), 25th (2), and 40th (3) positions. (b) The governing functions learned by Neural ODEs at the 10th (1), 25th (2), and 40th (3) positions.}
      \label{fig:Schrödinger Equation Model independent}
\end{figure}

Fig.~\ref{fig:Schrödinger Equation Model independent}(a) shows that  at the 10th, 25th, and 40th positions, the two latent representations now correspond to the wave function and its derivative, decoupled from each other:
\begin{equation}
     h_1(x) = a_1 \cdot \psi(x) + b_1, \quad h_2 = a_2 \cdot \frac{d\psi(x)}{dx} + b_2\tag{E.5}.
\end{equation}
Then, by calculating the derivatives with respect to the position \( x \), we can obtain:
\begin{equation}
    {f} =\frac{dh_2(x)}{dx} = \frac{a_2}{a_1}\cdot [V(x)\cdot h_1(x)] - \frac{a_2\cdot b_1}{a_1} \cdot [V(x)]\tag{E.6}.
\end{equation}
In Fig.~\ref{fig:Schrödinger Equation Model independent}(b), the governing functions learned by Neural ODEs  at the 10th, 25th, and 40th positions (z-axis) are plotted against \(V(x) \cdot h_1(x)\) ($x$-axis) and \(V(x)\) ($y$-axis). The plot clearly reveals a linear relationship among these variables, indicating that Neural ODEs have successfully captured the underlying equation.

\subsection{Pauli's spin-magnetic formulation}

We now attempt to decouple the four physical concepts previously identified in the main text. We assume that the true equations can be organized as two second-order equations. Equivalently, we can set:
\begin{equation}
    \frac{dh_{1} (x)}{dx} = h_{2}(x), \quad \frac{dh_{2}(x)}{dx} = {f}_1(h_{1}(x), h_{2}(x),h_{3}(x), h_{4}(x), V(x);\zeta_1)\tag{E.7}.
\end{equation}
\begin{equation}
    \frac{dh_{3}(x)}{dx} = h_{4}(x), \quad \frac{dh_{4}(x)}{dx} = {f}_2(h_{1}(x), h_{2}(x),h_{3}(x), h_{4}(x), V(x);\zeta_2)\tag{E.8}.
\end{equation}
Here \( h_1 \), \( h_2 \), \( h_3 \) and \( h_4 \) are the four latent states.

\begin{figure}[!ht]
    \centering
    \captionsetup{font=footnotesize, labelfont=bf}
\begin{minipage}[b]{0.48\textwidth}
        \centering
        \includegraphics[width=\textwidth]{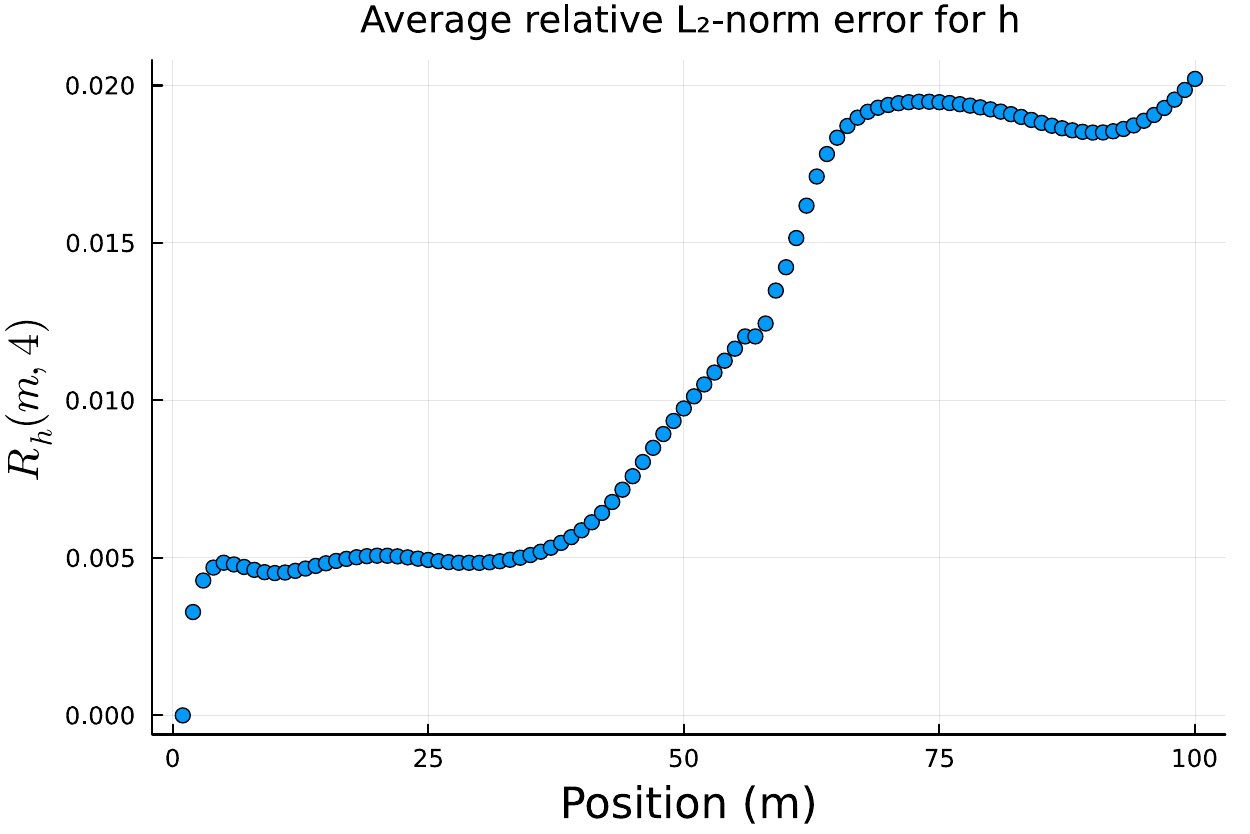}
        \subcaption{} 
    \end{minipage}
    \begin{minipage}[b]{0.48\textwidth}
        \centering
        \includegraphics[width=\textwidth]{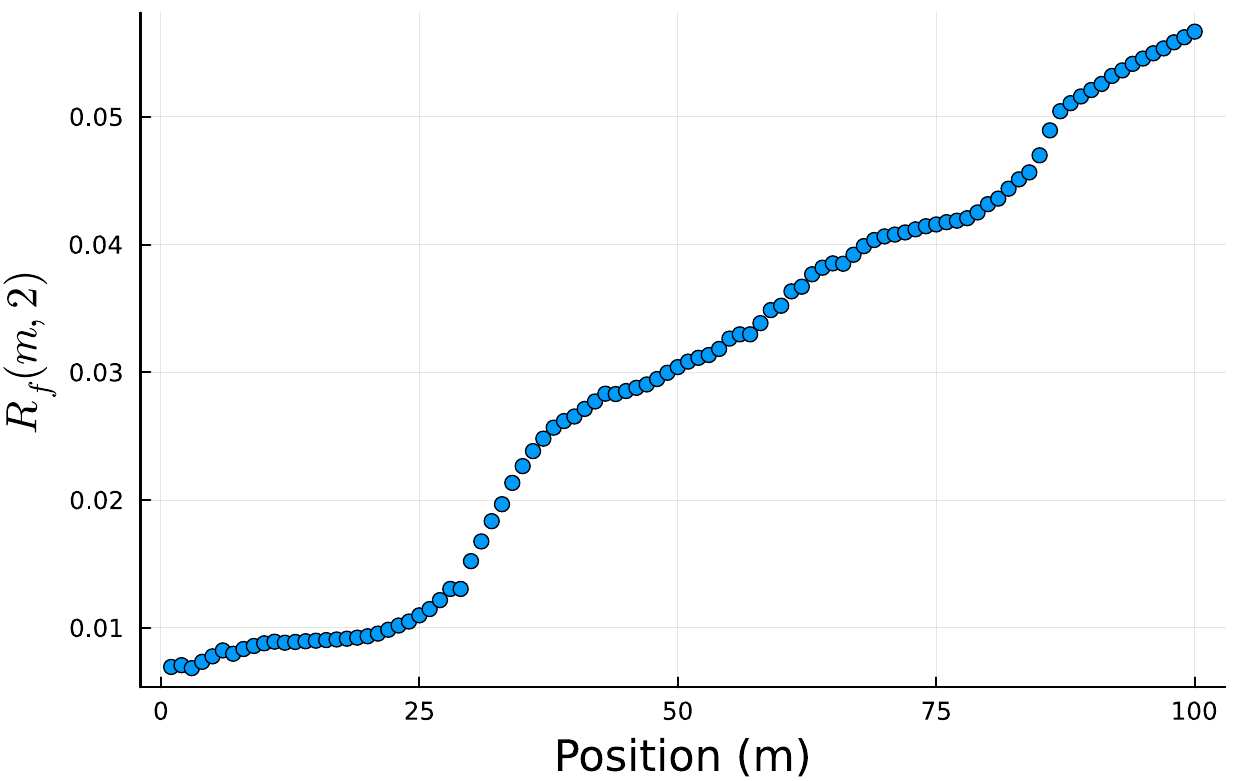}
        \subcaption{} 
    \end{minipage}
    
    \caption{The Pauli system with the second-order assumption. (a) Relative error metric between the learned latent representations and the linear combinations of the four physical concepts. 
(b) Relative error metric between the learned governing functions and the linear combinations of \({d\psi_1(x)/dx}\), \({d^2 \psi_1(x)}/{dx^2}\), \({d\psi_2(x)/dx}\) and \({d^2 \psi_2(x)}/{dx^2}\).}
    \label{fig:Pauli Equation Model independent}
\end{figure}

By comparing Fig.~\ref{fig:Pauli Equation Model independent} with Fig.~\ref{fig:Pauli Equation Model}(d) and Fig.~\ref{fig:Pauli Equation Model}(e), we observe that incorporating second-order information slightly improves the accuracy of equations identification; however, the learned latent representations remain linear combinations of four physical concepts.

\end{document}